\documentclass[conference]{IEEEtran}
\IEEEoverridecommandlockouts
\usepackage{cite}
\usepackage{amsmath,amssymb,amsfonts}
\usepackage{algorithm}
\usepackage{algpseudocode}
\usepackage{graphicx}
\usepackage{textcomp}
\usepackage{xcolor}
\usepackage{multirow}
\usepackage{subfig}
\usepackage{url}
\usepackage{bm}
\usepackage{array}

\newcolumntype{C}[1]{>{\centering\arraybackslash}p{#1}}

\def\BibTeX{{\rm B\kern-.05em{\sc i\kern-.025em b}\kern-.08em
    T\kern-.1667em\lower.7ex\hbox{E}\kern-.125emX}}

\begin{document}

\makeatletter
\newcommand{\linebreakand}{%
  \end{@IEEEauthorhalign}
  \hfill\mbox{}\par
  \mbox{}\hfill\begin{@IEEEauthorhalign}
}
\makeatother

\title{
Attention, Distillation, and Tabularization: Towards Practical Neural Network-Based Prefetching
}

\author{\IEEEauthorblockN{} 
%\author{\IEEEauthorblockN{Neelesh Gupta} 
\IEEEauthorblockA{\textit{} \\
 \\
}
}
\author{
\IEEEauthorblockN{Pengmiao Zhang\IEEEauthorrefmark{1}}
%\IEEEauthorblockN{Pengmiao Zhang}
\IEEEauthorblockA{\textit{University of Southern California} \\
Los Angeles, USA \\
pengmiao@usc.edu}
\and
\IEEEauthorblockN{Neelesh Gupta\IEEEauthorrefmark{1}} 
%\author{\IEEEauthorblockN{Neelesh Gupta} 
\IEEEauthorblockA{\textit{University of Southern California} \\
Los Angeles, USA \\
neeleshg@usc.edu}
\linebreakand

\IEEEauthorblockN{Rajgopal Kannan}
\IEEEauthorblockA{\textit{DEVCOM Army Research Lab}\\
Los Angeles, USA \\
rajgopal.kannan.civ@army.mil}
\and
\IEEEauthorblockN{Viktor K. Prasanna}
\IEEEauthorblockA{\textit{University of Southern California} \\
Los Angeles, USA \\
prasanna@usc.edu}
}

\newcommand{\ourwork}{DART}

\newcommand\blfootnote[1]{%
  \begingroup
  \renewcommand\thefootnote{}\footnote{#1}%
  \addtocounter{footnote}{-1}%
  \endgroup
}

\maketitle

\begingroup\renewcommand\thefootnote{* }
\footnotetext{These authors contributed equally.}
\endgroup

\begin{abstract}

Attention-based Neural Networks (NN) have demonstrated their effectiveness in accurate memory access prediction, an essential step in data prefetching. 
However, the substantial computational overheads associated with these models result in high inference latency, limiting their feasibility as practical prefetchers. 
To close the gap, we propose a new approach based on \textit{tabularization} that significantly reduces model complexity and inference latency without sacrificing prediction accuracy. Our novel tabularization methodology takes input as a distilled, yet highly accurate  attention-based model for memory access prediction and efficiently converts its expensive matrix multiplications into a hierarchy of fast table lookups. 
As an exemplar of the above approach, we develop \ourwork, a prefetcher comprised of a simple hierarchy of tables.
With a modest 0.09 drop in F1-score, \ourwork~reduces 99.99\% of arithmetic operations from the original attention-based model and 91.83\% from the distilled model.
\ourwork~accelerates the large model inference by 170$\times$ and the distilled model by 9.4$\times$.
\ourwork~has comparable latency and storage costs as state-of-the-art rule-based prefetcher BO but surpasses it by 6.1\% in IPC improvement.
\ourwork~outperforms state-of-the-art NN-based prefetchers TransFetch by 33.1\% and Voyager by 37.2\% in terms of IPC improvement, primarily due to its low prefetching latency.

\end{abstract}


\section{Introduction}

Data prefetching is a fundamental technique used in modern computing systems to bridge the latency gap between CPU cores and memory subsystems, thereby reducing overall program execution time~\cite{vander1997caches,carvalho2002gap}. A data prefetcher works by predicting which data will be needed in the near future and fetching it from memory in advance. This helps the processor avoid waiting for the data to be fetched from memory, significantly improving performance~\cite{dubois2012parallel}.

Existing table-based prefetchers have offered practical solutions using simple and fast table look-ups, relying on heuristic rules, such as spatial and temporal locality, to anticipate future memory accesses~\cite{kumar1998exploiting,mittal2016survey}. 
Best-Offset prefetcher (BO)~\cite{michaud2016best} employs a recent request table to record memory accesses and triggers prefetch requests. It updates scores of a list of offsets within a spatial range and predicts the offset with the highest score. Irregular Stream Buffer (ISB)~\cite{jain2013linearizing} devises table-based structures to track the most recent addresses and their corresponding program counters (PCs). Nonetheless, hampered by their reliance on heuristic and pre-established rules, these prefetchers exhibit limited adaptability and generalizability. They struggle to effectively discern intricate and latent patterns~\cite{zhang2022resemble,zhang2022fine,zhang2023phases}.
%, and their ability to accommodate the randomness stemming from multi-core parallel requests remains constrained~\cite{}.

On the other hand, machine learning algorithms, especially Neural Networks (NN), have showcased remarkable accuracy in predicting memory accesses. LSTM (Long Short-Term Memory)\cite{hochreiter1997long} is widely-used for memory access prediction~\cite{srivastava2019predicting,srivastava2020memmap,zhang2020raop,zhang2021c,shi2021hierarchical,hashemi2018learning} due to its proficiency in sequence modeling, but its recurrent architecture poses challenges for parallelization, resulting in high inference latency and low practicality. Attention-based models~\cite{vaswani2017attention} have demonstrated success in memory access prediction, achieving state-of-the-art performance and showing potential for practical hardware prefetcher deployment due to their high parallelism~\cite{zhang2022fine,zhang2023phases,zhang2022sharp}.

The central challenge in deploying NN-based prefetchers lies in their significant computational demands. 
Firstly, the quantity of parameters in well-performing NN models impose tremendous system resource requirements, potentially demanding more resources than the application that it was set out to prefetch for. 
Secondly, these models' complex nature leads to slow inference, causing untimely data prefetching. 
Lastly, the sheer volume of arithmetic operations inherent to NNs consumes both substantial resources and energy. 
{\color{black} Though existing works use pruning~\cite{liang2021pruning}, quantization~\cite{han2015deep}, and hardware parallel implementations~\cite{kyle2023mevit} to accelerate NN inference, these techniques still require a large number of matrix multiplications during inference.}

{\color{black} In light of this challenge, motivated by the success of table-based prefetchers, we propose a novel approach that transfers knowledge from a large attention-based neural network to a compact hierarchy of tables for more practical NN-based data prefetching, eliminating all matrix multiplications in model inference. }
The key steps of the proposed approach include: 1) training of a large attention-based model for memory access prediction, 2) reducing model complexity via knowledge distillation, and 3) implementing \textit{tabularization} to transform the distilled model into tables.

In developing the above approach, we tackle several key sub-problems that emerge. 1) How to map layers of neural networks with various operations to table lookups? We design kernels that not only store the precomputed matrix multiplications, but also manage challenges including bias incorporation, table size constraints, and the integration of activation functions between operations. 2) How to reduce the critical path of a model to satisfy a given latency constraint? We quantitatively analyze the table-based model’s inference latency and compress the NN model using a novel knowledge distillation approach for multi-label classification. 3) How to address the error accumulation when mapping multiple NN layers to tables? We propose a novel fine-tuning approach that trains the table to imitate the NN layer output rather than merely approximating dot products between layer inputs and weights. In this way, we mitigate performance degradation due to the approximation as the number of layers increases.

We develop a prefetcher~\ourwork~(\underline{D}istilling \underline{A}ttention-based neu\underline{R}al network to \underline{T}ables) as an exemplar of this approach. 
The predictor in~\ourwork~is simple and elegant: a hierarchy of tables acquired from the above steps.
\ourwork~couples the practicality inherent in table-based prefetchers with the precision offered by NN predictors. We summarize our main contributions as follows:

\begin{itemize}
    \item We propose a novel approach to transfer knowledge from an attention-based NN to a hierarchy of tables for the practical implementation of an NN-based prefetcher. 
    \item We design tabularization kernels for mapping the key operations in attention-based NN to table look-ups. Using the kernels, we can convert an arbitrary attention-based NN to a set of hierarchical tables.
    \item We develop a prefetcher~\ourwork~to exemplify our approach. We develop a table configurator for~\ourwork~to meet prefetcher design constraints.
    \item We propose a novel layer fine-tuning algorithm to mitigate the error accumulation problem when mapping more layers to tables.
    \item We evaluate the prediction performance of~\ourwork. With a 0.09 drop of F1-score, DART reduces 99.99\% of arithmetic operations from the large attention-based model and 91.83\% from the distilled model, accelerating the two models by 170$\times$ and 9.4$\times$, respectively.
    \item We evaluate the prefetching performance of~\ourwork~on a variety of workloads from SPEC 2006 and SPEC 2017. \ourwork~achieves a 37.6\% IPC improvement, outperforming state-of-the-art rule-based prefetcher BO by 6.1\% with comparable latency and storage costs. It also surpasses state-of-the-art NN-based prefetchers TransFetch and Voyager by 33.1\% and 37.2\%, respectively.
\end{itemize}

\section{Background}

\subsection{Attention}
\label{sec:back-attn}

The attention mechanism excels in memory access prediction due to its high accuracy, adaptability, and parallelizability~\cite{vaswani2017attention,zhang2022fine}. 
An attention-based model comprises two main components: a feed-forward network based on linear operations and multi-head self-attention based on dot-product attention. By crafting kernels to convert the fundamental linear and attention operations into table look-ups, we can tabularize any given attention-based model using the kernels.

\noindent\textbf{Feed-Forward Network (FFN).} An FFN is a stack of two linear operations along with a non-linear activation function:
\begin{equation}
\label{eq:linear}
\operatorname{Linear}\left( \bm{X}\right)=\bm{WX} + \bm{B}
\end{equation}
\begin{equation}
\label{eq:ffn}
\operatorname{FFN}(\bm{X})=\operatorname{Linear_O}\left( \max \left(0, \operatorname{Linear_H}\left( \bm{X}\right) \right) \right)
\end{equation}
where input $\mathbf{X} \in \mathbb{R}^{D_I \times T}$, weight $\bm{W}\in \mathbb{R}^{D_O \times D_I}$, and bias $\bm{B}\in \mathbb{R}^{D_O \times T}$. $T$, $D_I$, and $D_O$ are the input sequence length, input dimension, and output dimension, respectively. $\operatorname{Linear_H}$ and $\operatorname{Linear_O}$ are the hidden and output linear layers.

\noindent\textbf{Multi-Headed Self-Attention (MSA).} Given three distinct matrices: $\bm{Q}$, $\bm{K}$, $\bm{V} \in \mathbb{R}^{T \times D_{k}}$  the scaled dot-product attention is defined as:
\begin{equation}
\label{eq:attn}
\operatorname{Attention}(\bm{Q}, \bm{K}, \bm{V})=\operatorname{Softmax}\left(\frac{\bm{Q} \bm{K}^{\top}}{\sqrt{D_{k}}}\right) \bm{V}
\end{equation}
where $D_{k}$ is the dimension of $\bm{K}$. 
Consider the Attention input matrices are projected from the same input matrix though projection matrices \(\bm{W}_{i}^{{Q}}, \bm{W}_{i}^{{K}}, \bm{W}_{i}^{{V}} \in \mathbb{R}^{D \times D_{h}}\), \(h\) attention operations are concatenated as multiple heads, MSA is:
\begin{equation}
\begin{aligned}
\label{eq:msa}
\operatorname{MSA}(\bm{Q}, \bm{K}, \bm{V}) =\operatorname{Concat}\left(\operatorname{head}_{1}, \ldots, \text {head}_{\mathrm{h}}\right) \bm{W}^{O} \\
\text{head}_{\mathrm{i}} =\operatorname{Attention}\left(\bm{Q} \bm{W}_{i}^{{Q}}, \bm{K} \bm{W}_{i}^{{K}}, \bm{V} \bm{W}_{i}^{{V}}\right)
\end{aligned}
\end{equation}
where \(\bm{W}^O \in \mathbb{R}^{hD_{v}\times D}\) is the output layer weights that combines the results from each head, \(i\) is the index for each head, $D$ is the hidden dimension, \(D_v\) is the dimension of \(\bm{V}\), and \(D_h = D/h\) is the dimension of each head.

\subsection{Product Quantization}
\label{sec:back-pq}

Our tabularization approach is based on Product Quantization (PQ) \cite{jegou2010product}, a classic algorithm for approximating and accelerating vectors inner products through quantization and precomputation. Given a vector $\bm{a}\in\mathbb{R}^{D}$ that is drawn from a training set $\Tilde{\bm{A}}\in\mathbb{R}^{N\times D}$ with $N$ samples, and a fixed weight vector $\bm{b}\in\mathbb{R}^{D}$, PQ generates a quantized approximation $\hat{\bm{a}}$ such that $\hat{\bm{a}}^\top\bm{b}\approx\bm{a}^\top\bm{b} $. 
To quantize $\bm{a}$, the $D$-dimension vector is split to $C$ subspaces, each of dimension $V$. Within each subspace, $K$ prototypes are learned as the quantized subvectors for the subspace. Since $\hat{\bm{a}}$ is quantized and $\bm{b}$ is fixed, $\hat{\bm{a}}^\top\bm{b}$ can be precomputed and reused in a query. Figure \ref{fig:pq} shows an overview of PQ and the detailed process is introduced below.

\begin{figure}[h]
\centering
\subfloat[For product quantization training, prototypes within each subspace are learned, and dot products are precomputed to store in a table.]
{\includegraphics[width=\linewidth]{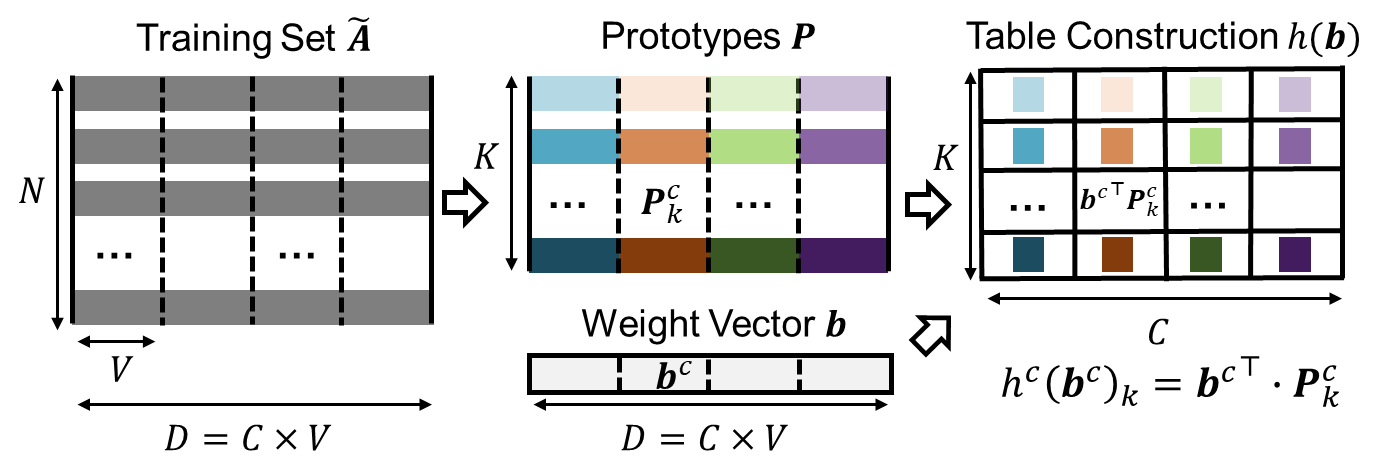}}\label{fig:pq_train}
\newline
\subfloat[For product quantization query, the input vector is encoded to find its nearest prototype index, retrieving it from the constructed table to aggregate and yield the final result.]
{\includegraphics[width=\linewidth]{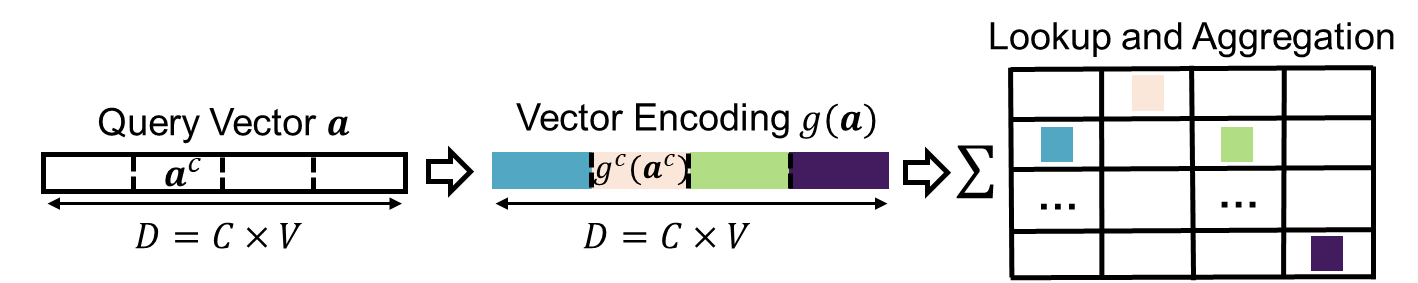}\label{fig:pq_query}}
\caption{Training and query of product quantization.}
\label{fig:pq}
\end{figure}

\subsubsection{Training} PQ training includes two main steps: learning the prototypes for each subsection and constructing the table which stores the precomputed results.

\noindent\textbf{Prototype Learning ($p$)}: Let $\Tilde{\bm{A}^c}\in\mathbb{R}^{N\times V}$ be subvectors in the $c$-th subspace from the training set $\Tilde{\bm{A}}$, the goal of this step is to learn $K$ prototypes $\bm{P}_k^c$ by minimizing the distance between each sub-vector $\Tilde{\bm{A}^c}$ and its nearest corresponding prototype $\bm{P}_k^c$ where $k$ signifies the indices of the prototypes within subspace $c$. The process is formulated in Equation \ref{eq:pq-1}.
\begin{equation}
\label{eq:pq-1}
p^c(\Tilde{\bm{A}}) \triangleq \underset{P}{\arg \min } \sum_c \sum_i\left\|\Tilde{\bm{A}}_i^c-\bm{P}_k^c\right\|^2
\end{equation}

\noindent\textbf{Table Construction ($h$)}:
Subsequently, a table is generated by computing and storing the dot product of each subspace's prototypes $\bm{P}_k^c$ and the weight vector $\bm{b}^c$ (the $c$-th subspace of $\bm{b}$). As detailed in Equation \ref{eq:pq-2}, the function $h^{c}(\bm{b})_k$ describes the entry that resides at index $(c,k)$ in the table.
\begin{equation}
\label{eq:pq-2}
h^{c}(\bm{b})_k \triangleq {\bm{b}^c}^\top \cdot \bm{P}_{k}^{c}
\end{equation}

\subsubsection{Query}
The query process avoids the multiplication operations in dot product by encoding the query vector to the closest prototype, looking up the precomputed results from the table, and perform aggregation. 

\noindent\textbf{Vector Encoding ($g$)}: Given the query vector $\bm{a}$, $g^c(\bm{a})$ identifies the closest prototype $\bm{P}^c_k$ in each subspace $c$ by finding index $k$ minimizing distance to $\bm{a}^c$, as in Equation \ref{eq:pq-3}. The outcome is a set of indices that represent the encoding of $\bm{a}$ using prototypes.
\begin{equation}
\label{eq:pq-3}
g^c\left(\bm{a}\right) \triangleq \underset{k}{\arg \min }\left\|\bm{a}^c-\bm{P}_k^c\right\|^2
\end{equation}
%The outcome is a set of indices that represent the encoding of $\bm{a}$ using prototypes, compressing it for quick retrieval.

\noindent\textbf{Table Lookup and Aggregation}: 
The dot products $\bm{a}^\top\bm{b}$ can be approximated by looking up the precomputed values through the encoded indices, and aggregating the subspaces through an aggregation function $f(\cdot)$, as shown in Equation~\ref{eq:pq-4}.
\begin{equation}
\label{eq:pq-4}
f(\bm{a},\bm{b})=\sum_{c} h^{c}(\bm{b})_k, k=g^{c}(\bm{a})
\end{equation}

The query avoids the dot product operation in $\bm{a}^\top\bm{b}$ by approximating the result through table look-ups. We use locality sensitive hashing~\cite{blalock2021multiplying} for encoding and use parallel summation for aggregation, the complexity is significantly lower than the dot product, especially for large dimension vectors.

\section{Related Works}

\subsection{ML for Data Prefetching}

Although hardware prefetching is well-researched, applying Machine Learning (ML) has seen a surge due to stagnation in existing approaches~\cite{islam2016prefetching,choi2021survey}.
Prior works have exhausted Long Short-Term Memory (LSTM) models as initial gauges to assess the potential of ML-based prefetching improvements~\cite{hashemi2018learning,srivastava2019predicting,srivastava2020memmap}.
%LSTMs, with their ability to adapt to a variety of classification tasks, have served as a backbone in exploring the trade-offs in implementing ML models for prefetching tasks due to their substantial inference and storage overheads .
Recent ML-based prefetchers exploit attention mechanisms to emphasize salient patterns, setting unparalleled benchmarks in prediction performance \cite{shi2021hierarchical,zhang2022fine,zhang2022a2p,yang2023gl}.
Other ML-based prefetching optimizations have tested logistic regression, decision trees, random forests, and reinforcement learning for both cache-level memory access prediction and the selection of existing hardware prefetchers \cite{Rahman2015-xk, peled2015semantic, zhang2022resemble, eris2022puppeteer, alcorta2023lightweight}.
Despite high performance in memory access prediction, previous approaches neglect the practicalities essential for real-world deployment of ML-based prefetchers, most notably latency and storage considerations \cite{mohapatra2023drishyam}.
% Our work bridges to gap between SOTA performance and practicality.

\subsection{Neural Network Acceleration and Approximation}

%Addressing the complexity of s demands innovative acceleration techniques \cite{cheng2017survey,chen2020survey}.
Various techniques have been explored to reduce Neural Network (NN) complexity and accelerate inference. 
Model compression techniques, such as parameter pruning~\cite{liang2021pruning,liu2019rethinking}, quantization \cite{han2015deep,nagel2021white}, low-rank factorization \cite{sainath2013low}, and knowledge distillation \cite{hinton2015distilling, gou2021knowledge}, target redundancy and model size to boost performance.
In hardware, %ReRAM optimizes memory with high storage density \cite{liu2016memristor,qiao2018atomlayer}, 
TPUs with systolic arrays allow for tensor processing demands at scale \cite{jouppi2017datacenter}, and FPGAs offer parallel acceleration tailored for CNNs and GNNs \cite{abdelouahab2018accelerating,abikaram2022gengnn}.
While these methods reduce model complexity or accelerate computation, they still require matrix multiplications during inference.
Approximate Matrix Multiplication techniques replace matrix multiplications using hashing and averaging \cite{blalock2021multiplying,francis2022practical}, while shift-and-add strategies aim to render NNs multiplication-free, optimizing power efficiency \cite{you2020shiftaddnet,elhoushi2021deepshift}. 
Current methods either concentrate on the final NN layer or simply alter matrix multiplication. In contrast, our paper introduces a novel approach that transforms the entire NN into a customizable table hierarchy while maintaining performance. To the best of our knowledge, our work represents the first instance of a tabularized NN-based prefetcher with a comprehensive design methodology and training scheme.

\section{Overview}

Our objective is to bridge the gap between traditional table-based prefetchers and NN-based prefetchers by reconfiguring the neural network inference process into a rapid and efficient table look-up mechanism, {\color{black} eliminating the need for matrix multiplications during model inference.}

\subsection{Problem Definition}

Given the sequences of $T$ historical physical block addresses $\bm{A}_t = \{a_{1}, a_{2}, ..., a_T\}$ and historical program counters $\bm{PC}_t = \{pc_{1}, pc_{2}, ..., pc_T\}$ at time $t$, the input for the prediction models is denoted as $\bm{X}_t = \{(a_1, pc_1), (a_2, pc_2) ..., (a_T, pc_T)\}$. The future $K$ block addresses are represented by $\bm{Y}_t=\{y_1, y_2, ..., y_K\}$. Prefetcher design constraints include a latency constraint $\tau$ and a storage constraint $s$. Our objective is to construct a tabular model $\mathcal{T}(\bm{X}_t; \boldsymbol{\theta})$ that learns the mapping from $\bm{X}_t$ to $\bm{Y}_t$, such that:

\begin{equation}
\begin{aligned}
    \mathcal{L}(\mathcal{T})<\tau, \mathcal{S}(\mathcal{T})< s
    %\frac{1}{n} \sum_{i=1}^{n} (y_{i} - y_{i}')^2 < \epsilon(\tau,s) 
%    T(\mathcal{M'(\mathbf{x}; \boldsymbol{\theta'})})<\tau
\end{aligned} 
\end{equation}
where $\mathcal{L}(\cdot)$ represents the model inference latency and $\mathcal{S}(\cdot)$ represents the model storage cost.

\subsection{Overall Approach}

We develop a three-fold approach to solve this problem as illustrated in Figure~\ref{fig:overview_approach}. The workflow consists of a preprocessing step and three key training steps: 1) {\bf Attention}: Training a high-performance prediction model, 2) {\bf Distillation}: Reducing model complexity to meet the prefetcher design constraints, and 3) {\bf Tabularization}: Converting the model to a hierarchy of tables to achieve fast model inference. 

\begin{figure}[h]
    \centering
    \includegraphics[width=\linewidth]{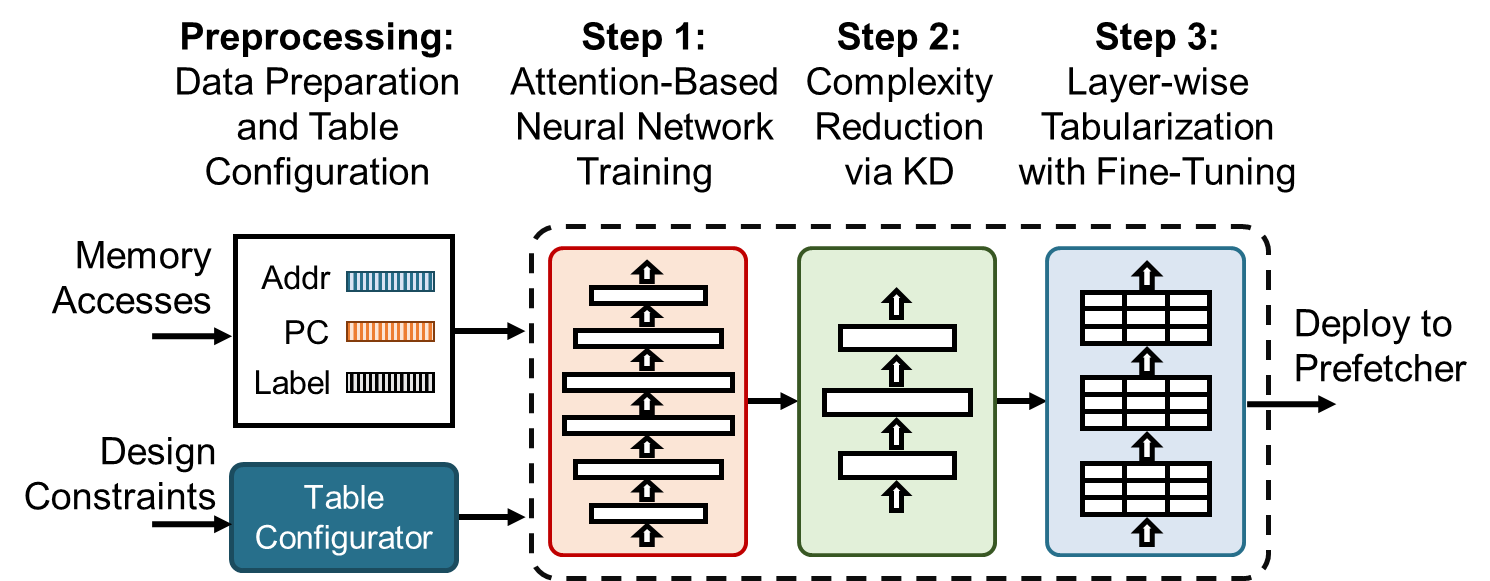 }
    \caption{Approach to constructing compact table-based memory access predictor by distilling knowledge from a trained attention-based neural network.}
    \label{fig:overview_approach}
\end{figure}

\noindent \textbf{Preprocessing: Data Preparation and Table Configurator.} 
To train the model, we first preprocess the data and set the configurations for both the neural network and table-based models based on design constraints $\tau$ and $s$. These constraints directly impact the table-based predictor's latency and storage, contingent on the NN's layers and dimension size. Using these parameters, a table configurator formulates the optimal attention-based and table-based model structures.

\noindent \textbf{Step 1: Attention-based Neural Network Training.}
With the prepossessed data and labels, we train a large and robust attention-based multi-label memory access prediction model $\bm{\hat{Y}}_t = \mathcal{M}(\bm{X}_t; \boldsymbol{\theta})$.
The aim of this step is to pursue exceptional prediction performance, regardless of any limitations imposed by design constraints or model complexity.

\noindent \textbf{Step 2: Complexity Reduction via Knowledge Distillation.}
The well-trained model $\mathcal{M}$ is large and typically cannot meet the prefetcher design constraints. Therefore, we need to reduce the model complexity by reducing the number of layers and dimensions. The adaptability of Knowledge Distillation (KD) \cite{hinton2015distilling} allows us to tailor the target model to meet specific constraints. 
Using the well-trained large model $\mathcal{M}$ as the teacher network, we employ KD to train a shallower and more compact attention-based student network $\mathcal{M}_s$.
Then, we adopt the predictor configuration generated by the table configurator to ensure the network can map to a valid table-based predictor that meets the prefetcher design constraints.

\noindent \textbf{Step 3: Layer-wise Tabularization with Fine-Tuning.} 
To make our NN-based prefetcher practical, we transform the KD compressed model $\mathcal{M}_s$ into a table-based structure $\mathcal{T}$ for quick look-up-based inferences. This is dubbed \textit{Tabularization}. We've crafted specialized kernels for linear and attention functions to facilitate this process for an arbitrary attention-based NN. With these kernels, we systematically convert our distilled model into organized tables, all while considering predefined design constraints. However, as the process progresses, cumulative errors between subsequent layers can worsen. To counteract this, we employ a layer fine-tuning method to adjust weights in light of the preceding tabular layer outcomes.

\subsection{DART}

As an exemplar of our proposed approach, we introduce~\ourwork, a prefetcher equipped with a table-based predictor derived through distilling knowledge from an attention-based memory access prediction model, as is shown in Figure~\ref{fig:overview_dart}.~\ourwork~is integrated into the last level cache (LLC), uses LLC memory accesses as input, predicts future memory accesses, and request prefetch to LLC. The prediction process is mainly table look-ups layer by layer with a minimal number of arithmetic operations. The model inference is fast and the model implementation is simple. The training of~\ourwork~follows the proposed approach, detailed in Section~\ref{sec:DART_Training}.

\begin{figure}[h]
    \centering
    \includegraphics[width=0.78\linewidth]{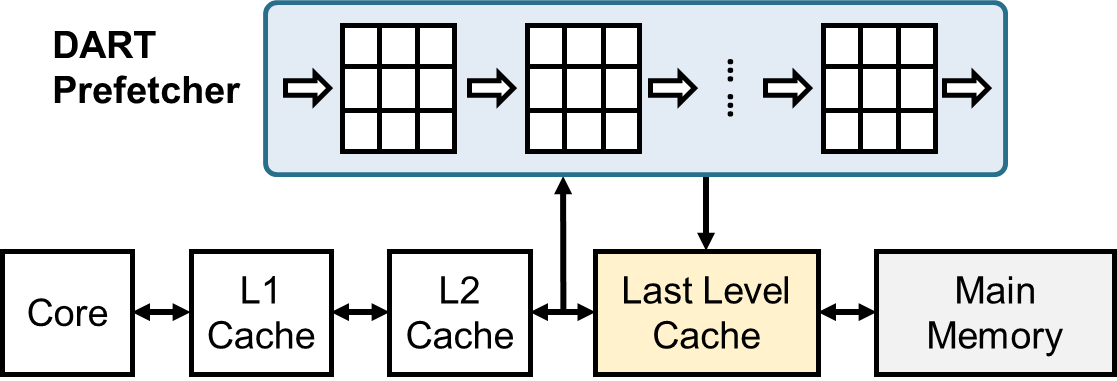 }
    \caption{Using the proposed approach, we present our LLC prefetcher~\ourwork~with a table-based predictor distilled from an attention-based memory access prediction model.}
    \label{fig:overview_dart}
\end{figure}

\section{Tabularization Kernels}
\label{sec:kernel}
We design tabularization kernels to convert the key operations in an attention-based NN into table look-ups.
These operations consist of linear operations and attention operations (Section~\ref{sec:back-attn}).
By designing these kernels, the inference of an arbitrary attention-based NN can be mapped to table look-ups.

The kernels are based on the technique of product quantization (PQ) in Section~\ref{sec:back-pq}, eliminating expensive matrix multiplication operations. Furthermore, we propose three optimizations for attention-based NN tabularization. First, to accommodate the $T$-length input sequence, we extend the PQ approach from two vectors to the product of matrices. Second, to further condense NN operations, we merge the bias addition operation and activation functions to the table construction and avoid these steps in the query. Third, to tabularize the attention operation without a fixed weight matrix, we generalize the approach to accommodate two variable matrix multiplications.
We outline the training and querying processes for the kernels.

\subsection{Linear Kernel} 

A linear operation in a NN is shown as in Equation~\ref{eq:linear}.
Assume the training set is $\bm{\Tilde{X}} \in \mathbb{R}^{N\times T \times D_I}$ with $N$ number of samples. Bias $\bm{B}$ in Equation~\ref{eq:linear} is a $T$ repeat of bias vector $\bm{b}$ adding to all time-step (sequence length) outputs. We define the number of subspaces as $C$ and number of prototypes  within a subspace as $K$ for tabularization. Figure~\ref{fig:kernel_linear} illustrates the linear kernel training and query process. 

\begin{figure}[h]
\centering
\subfloat[For linear kernel training, prototypes are learned from row vectors across samples and sequences, adding bias as a column to trim query operations.]
{\includegraphics[width=\linewidth]{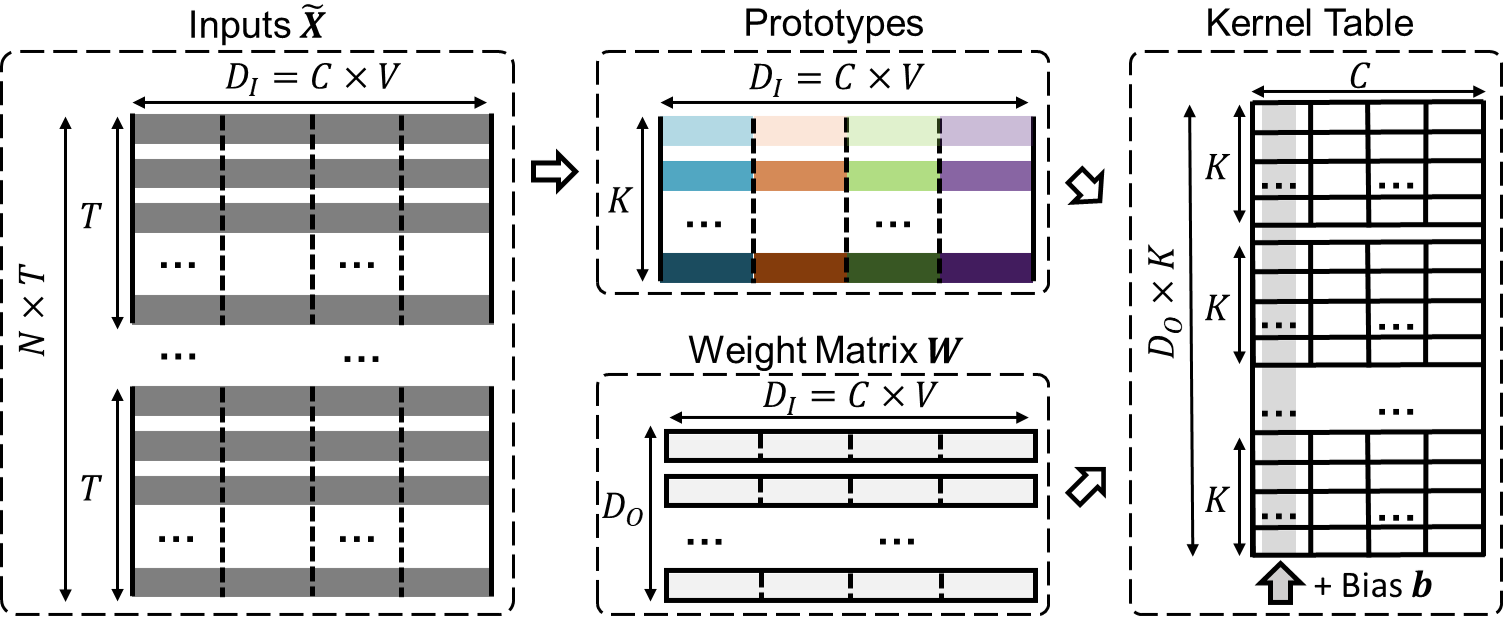}\label{fig:kernel_linear_training}}
\newline
% \subfloat[Linear kernel query eliminates matrix multiplication and bias addition. All $T$ dimensions perform query in parallel.]
\subfloat[For linear kernel query, the input vector is encoded and looked up in parallel across T dimensions.]
{\includegraphics[width=\linewidth]{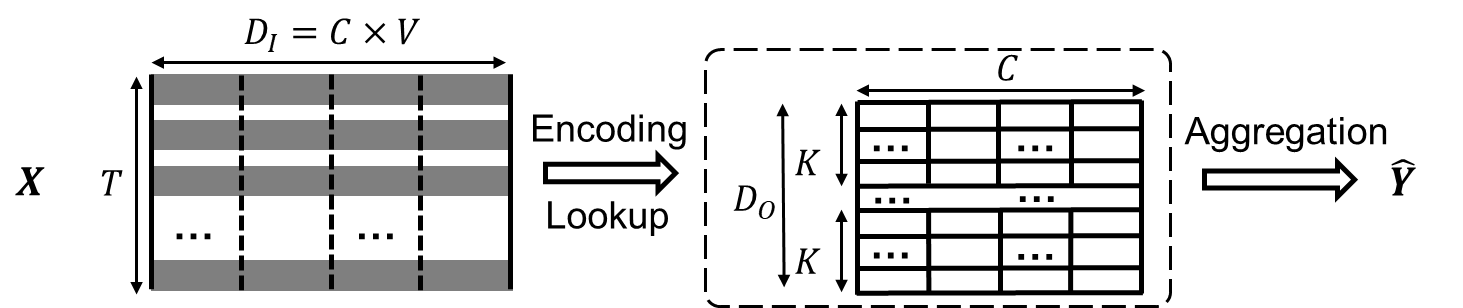}\label{fig:kernel_linear_query}}
\caption{Training and query of the linear kernel.}
\label{fig:kernel_linear}
\end{figure}

\begin{figure*}[h]
\centering
\subfloat[For attention kernel training, pairwise products of learned prototypes are stored for query reuse and doubly quantized to limit table size, integrating Softmax and scaling directly into the QKV table result.]
{\includegraphics[width=\linewidth]{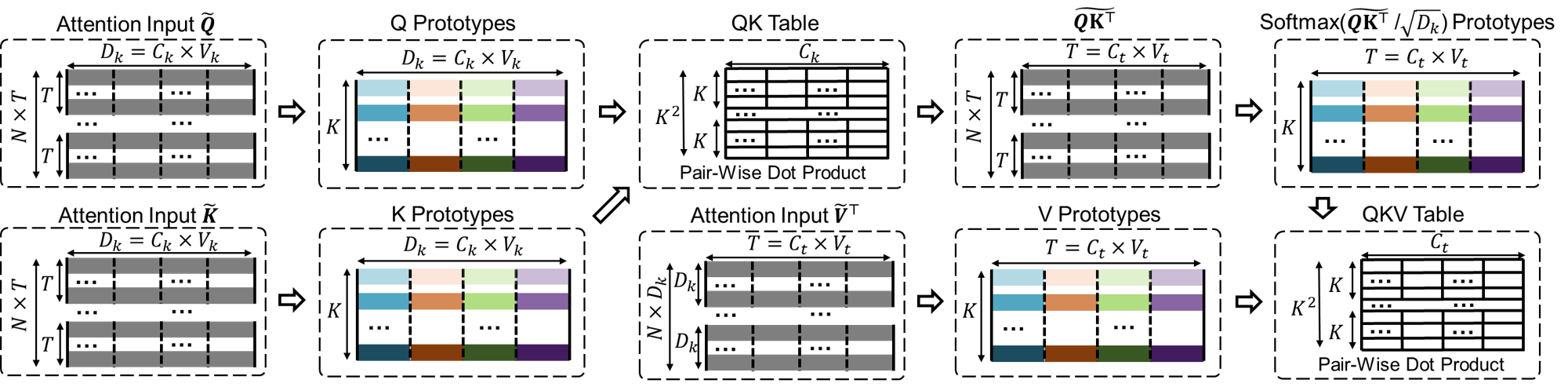}\label{fig:kernel_attn_training}}
\newline
\subfloat[For attention kernel query, matrix multiplication, scaling, and activation are replaced by two lookups: from $\bm{Q}$, $\bm{K}$ to $\hat{\bm{Q{K^\top}}}$, and $\hat{\bm{Q{K^\top}}}$, $\bm{V}$ to the result.]
{\includegraphics[width=\linewidth]{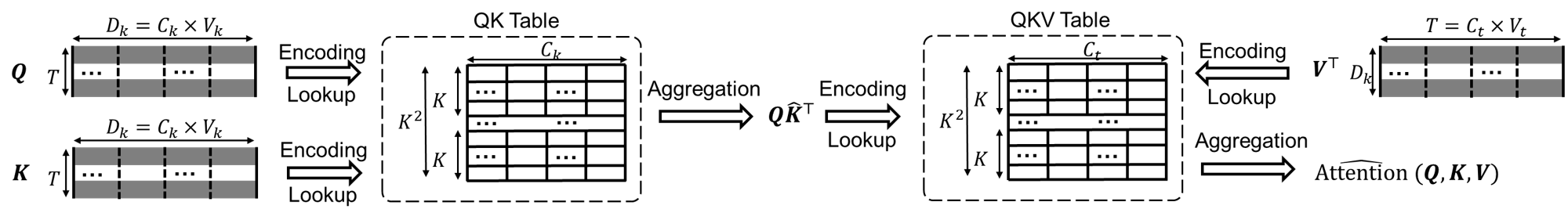}\label{fig:kernel_attn_query}}
\caption{Training and query of the attention kernel.
}
\label{fig:kernel_attn}
\end{figure*}

\subsubsection{Training}
Figure~\ref{fig:kernel_linear_training} illustrates the kernel training process. We reshape $\bm{\Tilde{X}}$ from $\mathbb{R}^{N\times T \times D_I}$ to $\bm{\Tilde{X}}_r \in \mathbb{R}^{NT \times D_I}$ to learn prototypes of dimension $D_I$. For each prototype $P^c_k$ in subspace $c$, a dot product between linear layer weights $W_o^c$ is computed and stored for reuse. The outcome populates table element $h^c_{o}(\bm{W})_k$. 
To merge bias addition, we reshape the bias to match the constructed table $\bm{b}_r \in \mathbb{R}^{D_O\times K \times C}$, repeating values in each $Do$ dimension $K$ times for the first $C$ subspace column and filling the rest with zeroes. $\bm{b}_r$ is then added to the learned table, equivalent to integrating the bias $\bm{b}$ into a single subspace dimension. Thus, during query aggregation, the bias will be automatically added to the final result. The formal expression for the table construction is: 
\begin{equation}
\label{eq:kernal_linear}
h_o^{c}(\bm{W})_k = {\bm{W}_o^c}^\top \cdot p^c({\bm{\Tilde{X}} _r})_k+\bm{b}_r
\end{equation}
%\noindent\textbf{Query}.

\subsubsection{Query} 
The query of a single row of $\bm{X}$ is similar to the basic PQ query. All $T$ row vectors in $\bm{X}$ are encoded and generate results for all $D_O$ output dimensions. These encoding, table look-ups, and aggregations are independent so they are embarrassingly parallel. 
By aggregating the results over $C$ subspaces, the output $\hat{\bm{Y}} \in \mathbb{R}^{T \times D_O}$is:
\begin{equation}
\hat{\bm{Y}}_{t,o} = \sum_{c} h_o^{c}(\bm{W})_k, k=g^{c}(\bm{X}_t)
\end{equation}

\subsection{Attention Kernel}

In Equation~\ref{eq:attn}, we show the attention mechanism.
Compared to linear operations, tabularizing attention processes proves more challenging.
Firstly, the absence of a fixed weight matrix renders the precompute processing in linear kernels impossible.
To counter this hurdle, we tabularize the pair-wise dot products.
Secondly, attention operations requires three matrix multiplications, which can potentially balloon the table depth to $K^3$ for $K$ prototypes, leading to high storage consumption.
Our solution involves a secondary quantization of the intermediate result to trim the table depth down to $2K^2$.
Lastly, beyond mere matrix multiplication, operations such as scaling and Softmax activation introduce added steps and latency.
We've streamlined this process by integrating the operations into prototypes in training, removing need for such operations during a query. 

\subsubsection{Training} 

Figure~\ref{fig:kernel_attn_training} shows the kernel training process to construct tables for attention operation. First, We train $K$ prototypes of the training data set $\bm{\Tilde{Q}}\in \mathbb{R}^{N\times T \times D_k}$ and $\bm{\Tilde{K}}\in \mathbb{R}^{N\times T \times D_k}$, where $N$ is the number of samples, $T$ is the sequence length, and $D_k$ is the dimension size. Within each subspace of $C_k$, we compute and store the pair-wise dot product of the Q prototypes and K prototypes, generating a $K^2$ depth table with $C_k$ width, denoted as QK table. Second, based on the training set $\bm{\Tilde{Q}}$ and $\bm{\Tilde{K}}$ and the trained QK table, we can generate the corresponding $\bm{\Tilde{QK^\top}} \in \mathbb{R}^{N\times T \times T}$. We perform a second quantization process to $\bm{\Tilde{QK^\top}}$ by training its $K$ prototypes. Third, we directly process the scaling of $\sqrt{D_k}$ and Softmax activation on the prototypes, then use this processed prototype to perform pair-wise dot product with the other input of attention $\bm{\Tilde{V}}\in \mathbb{R}^{N\times T \times D_k}$, generating a table at depth $K^2$ with width $C_t$ (the number of subspaces for $T$ dimension), denoted as QKV table. In summary, through two quantization steps and two pair-wise dot produce, the tabularization output is two tables: QK table at size $K^2\times C_k$ and a QKV table at size $K^2\times C_t$.
The tabularization process of QK table is:
\begin{equation}
%\label{eq:kernal_linear}
h^{c}(\bm{\Tilde{Q}},\bm{\Tilde{K}})_{i,j} = p^c({\bm{\Tilde{Q}}_r})_i \cdot p^c({\bm{\Tilde{K}} _r})_j
\end{equation}
where $\bm{\Tilde{Q}}_r$ and $\bm{\Tilde{K}} _r$ are reshaped input matrix with size $\mathbb{R}^{NT \times D_k}$, $i$ and $j$ are the index of prototypes. The output of the QK table through aggregation is as below:
\begin{equation}
\label{eq:kernel_h_qk}
\bm{\Tilde{QK^\top}}_t = \sum_{c} h^{c}(\bm{\Tilde{Q}}_t,\bm{\Tilde{K}}_t)_{i,j}, i=g^{c}(\bm{\Tilde{Q}}_t), j=g^{c}(\bm{\Tilde{K}}_t)
\end{equation}
The generated results from QK table along with the reshaped and transposed the attention input $\bm{\Tilde{V}}^\top \in _r\mathbb{R}^{ND_k \times T}$ performs pair-wise dot product and construct the table for lookup:
\begin{equation}
%\label{eq:kernal_linear}
h^{c}(\bm{\Tilde{QK^\top}},\bm{\Tilde{V}})_{i,j} =  \text{Sigmoid}(p^c({\bm{\Tilde{QK^\top}}})_i/\sqrt{D_k}) \cdot p^c({\bm{\Tilde{V}} _r})_j
\end{equation}
The two constructed tables are stored for query while the learned prototypes are not.

\subsubsection{Query} 
The query process of attention kernel is shown in Figure~\ref{fig:kernel_attn_query}. The query process consists of two critical steps of table look-ups. First, the input matrix $\bm{Q}$ and $\bm{K}$ is encoded and look-up for the dot product results in QK table as in Equation~\ref{eq:kernel_h_qk} and get estimated $\hat{\bm{Q{K^\top}}}$, Then $\hat{\bm{Q{K^\top}}}$ is encoded again and along with the encoded $V$ input to look up from the QKV table and aggregates for the final estimation of attention operation as below:
\begin{equation}
%\label{eq:kernal_linear}
\hat{\bm{Y}}_t  =\sum_{c} h^{c}(\hat{\bm{Q{K^\top}}},\bm{{V}})_{i,j} , i=g^{c}(\hat{\bm{Q{K^\top}}}_t), j=g^{c}(\bm{{V}}_t)
%\sum_{c} h^{c}(\bm{\Tilde{Q}}_t,\bm{\Tilde{K}}_t)_{i,j}, i=g^{c}(\bm{\Tilde{Q}}_t^c), j=g^{c}(\bm{\Tilde{K}}_t^c)
\end{equation}
The query of attention kernel avoids all matrix multiplications, scaling calculation, and activation operations.

%\newpage

%============
\subsection{Kernel Complexity}
\label{sec:kernel_complexity}
We analyze the kernel complexity by thoroughly examining the latency, storage, and the number of arithmetic operations associated with the proposed kernels. These factors will be instrumental in shaping the overall model design to align with prefetcher constraints.

\subsubsection{Latency ($\mathcal{L}$)} We assume fully parallel implementation. We use the locality sensitive hashing in~\cite{blalock2021multiplying} as the encoding function $g$, the latency is $\log(K)$ for $K$ prototypes. 

\noindent \textbf{Linear kernel latency.} 
The linear kernel latency consists of the latency for encoding subvectors $g$, the latency for table lookup $h$ and the latency for subspace aggregation $f$:
\begin{equation}
%\label{eq:linear_latency}
   \mathcal{L}_l(K,C) = \mathcal{L}_g+ \mathcal{L}_f+\mathcal{L}_h = \log(K) + \log(C) + 1
\end{equation}
\noindent \textbf{Attention kernel latency.} 
The attention kernel latency consists of the latency of the input encodings $g$ for $\bm{Q}, \bm{K},$ and $ \bm{V}$, the aggregation for QK Table, the encoding $g_{qk}$ for the approximated $\bm{QK^\top}$, and the final aggregation $f$:
\begin{equation}
%\label{eq:attn_latency}
\begin{aligned}
   &\mathcal{L}_a (K,C) = \mathcal{L}_{g_{i},g_{qk}}+ \mathcal{L}_{f_{qk}, f_{qkv}} +\mathcal{L}_{h_{qk},h_{qkv}}
   \\
   &= 2\log(K) + \log(C_k) +\log(C_t) + 2 \\
   &\Rightarrow2(\log(K)+\log(C)+1)\text{, if }  C=C_k=C_t
\end{aligned}
\end{equation}

\subsubsection{Storage ($\mathcal{S}$)} The storage cost of a kernel consists of the table entries and the vector encoding results. One index of the encoded prototypes takes storage cost at $log(K)$ bits. For the precomputed table entries, we denote the data bit-length as $d$. The prototypes do not need to be stored since we use the encoded indices to look-up from the table directly.

\noindent \textbf{Linear kernel storage.} Encoding of the input subvectors $g$ assigns them to $TC$ indices of prototypes. The total number of table entry is $D_OKC$ as shown in Figure~\ref{fig:kernel_linear} The linear kernel storage in bit is:
\begin{equation}
\label{eq:linear_storage}
\begin{aligned}
   &\mathcal{S}_l^d (T,D_O,K,C) = \mathcal{S}_g+ \mathcal{S}^d_h 
   = TC\log(K) + D_OKCd
\end{aligned}
\end{equation}

\noindent \textbf{Attention kernel storage.} As shown in Figure~\ref{fig:kernel_attn}, there are four encoding operations and two tables, the storage cost is:
\begin{equation}
\label{eq:attn_storage}
\begin{aligned}
   &\mathcal{S}_a^d (T,D_k,K,C) = \mathcal{S}_{g_q, g_k, g_{qk}, g_v}+ \mathcal{S}^d_{h_{qk}, h_{qkv}}\\
   &= (2TC_k+TC_t+D_kC_t)\log(K)+K^2(C_k+C_t)d \\
   &\Rightarrow (3T+D_k)C\log(K)+2K^2Cd\text{, if }  C=C_k=C_t
\end{aligned}
\end{equation}
where $g_q, g_k, g_v, g_{qk}$ is for the encoding of attention input matrices $\bm{Q},\bm{K},\bm{V}$ and intermediate result $\bm{QK^\top}$. $h_{qk}$ and $h_{qkv}$ are the QK table and the QKV table in Figure~\ref{fig:kernel_attn}.
%\newpage

\subsubsection{Arithmetic Operations ($\mathcal{A}$)}
We analyze the arithmetic operations for the kernels besides table look-ups. 

\noindent\textbf{Linear kernel arithmetic operations.} The operations consists of two parts, the encoding $g$ to get table indexes and the aggregation $f$ after table look-up to get the output results. 
\begin{equation}
\label{eq:linear_storage}
%\begin{aligned}
   \mathcal{A}_l = \mathcal{A}(g)+ \mathcal{A}(f) = TC\log(K)+TD_O\log(C)
%\end{aligned}
\end{equation}
\noindent\textbf{Attention kernel arithmetic operations.} There are four encoding process $g$ and two aggregation $f$ processes. 
\begin{equation}
\label{eq:attn_opslinear_storage}
\begin{aligned}
   \mathcal{A}_a &= \mathcal{A}(g_q, g_k, g_{qk}, g_v)+ \mathcal{A}(f_{qk}, f_{qkv})\\
   &=(2TC_k + TC_t+D_kC_t)\log(K) \\
   &~~~~+ T^2\log(C_k) + D_k^2\log(C_t)\\
   %   &=[2TC_k + (T+D_k)C_t]\log(K) + T\log(C_k) + D_k\log(C_t)\\
   &\Rightarrow (3T+D_k)C\log(K) + (T^2+D_k^2)\log(C),\\
   &~~~~\text{if }  C=C_k=C_t
   %= D_OC\log(K)+D_O\log(C)
\end{aligned}
\end{equation}

Using the designed kernels, we are able to accelerate the inference of a given attention-based neural network by converting its key operations to table look-ups. 

\section{Training of DART}
\label{sec:DART_Training}
In this section, we present the training scheme of DART, an exemplar of our approach to distill knowledge from an attention-based neural network to a hierarchy of tables towards practical prefetching. The key steps include data preparation (Section~\ref{sec:data_prep}), training of an attention-based model (Section~\ref{sec:approach_attn}), configuring table structure based on prefetcher constraints (Section~\ref{sec:tab_config}), reducing model complexity through knowledge distillation (Section~\ref{sec:approach_kd}), and layer-wise tabularization using the designed kernels (Section~\ref{sec:finetune}).

\subsection{Data Preparation}

\label{sec:data_prep}
We extract memory access trace from the last level cache. We process the extracted trace following TransFetch~\cite{zhang2022fine} to get the input sequences and labels for training the memory access prediction models and the tables.

\noindent\textbf{Segmented address input}. We dissect a block address into $S=\lceil\frac{p}{c}\rceil+1$ segments for $p$-bit page address and $c$-bit block index, each housing $c$ bits. This process maps a memory address to an $s$-dimensional vector, transforming a $T$-length sequence to a $T\times S$ matrix for attention-based model input.

\noindent \textbf{Delta bitmap labels}. We use a bitmap that each position indicates a delta between upcoming and current memory addresses. If a delta that appears in a near future (within a look-forward window) and falls in a range, the corresponding bit in the bitmap is set to 1. This approach enables the model to issue multiple predictions simultaneously.

\subsection{Attention-Based Neural Network Training}
\label{sec:approach_attn}
We train a large attention-based neural network for memory access prediction. This model is to pursue high prediction performance without taking the prefetcher design constraints into consideration.

\noindent\textbf{Model architecture}. 
We use the commonly used architecture, visualized in Figure~\ref{fig:vit_map}, employing segmented addresses as inputs into the Transformer encoder layers, where each encoder layer consists of MSA mechanisms and FFNs. Following the encoder layers, the processed data is forwarded to a classification head that generates a delta bitmap, indicating multiple predicted address deltas to the current address.

\begin{figure}[h!]
    \centering
    \includegraphics[width=0.95\linewidth]{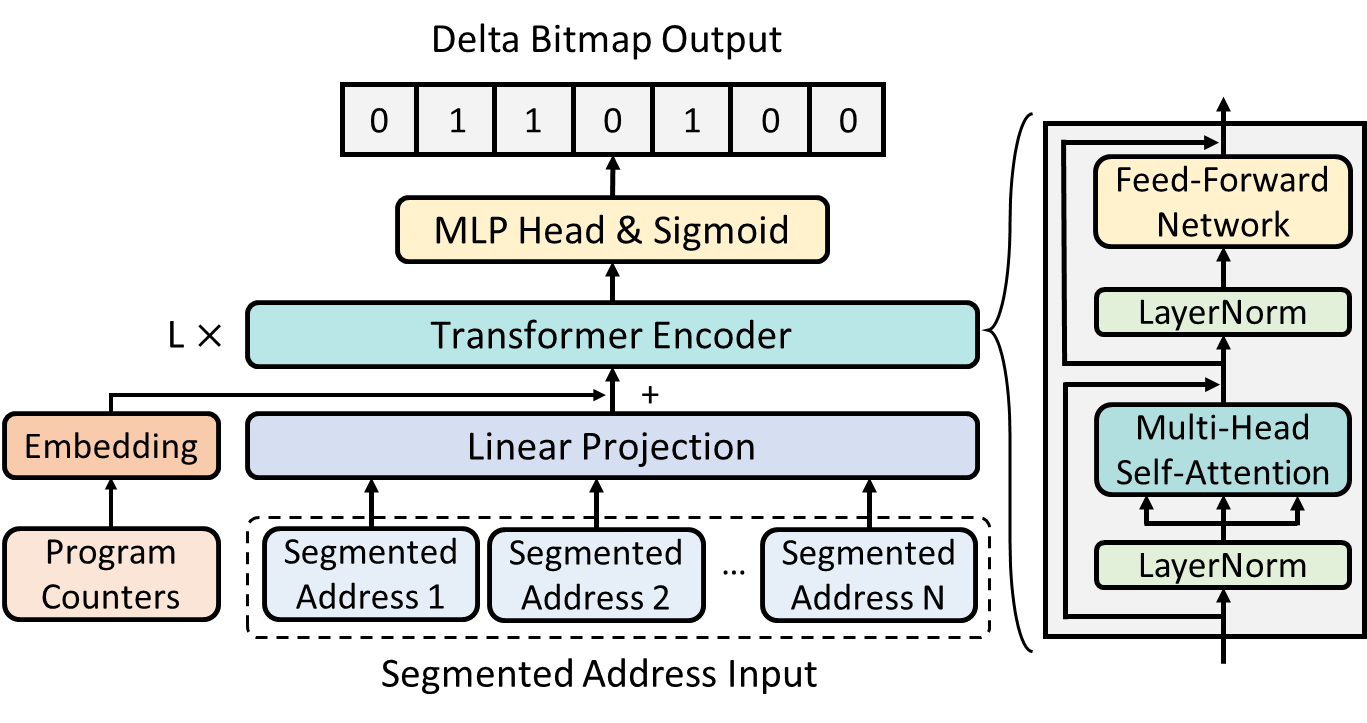}
    \caption{Attention-based network for memory access prediction.}
    \label{fig:vit_map}
\end{figure}

\noindent\textbf{Loss function}. We train the model as multi-label classification using Binary Cross-Entropy (BCE) loss.

\subsection{Table Configuration}
\label{sec:tab_config}
{\color{black}Given design constraints on latency $\tau$ and storage $s$, we devise a table configurator to determine the structure of the final hierarchy of tables. 
We analyze the entire model's latency and storage, then present the table configurator mechanism.}

\begin{table}[h]
  \caption{Notations for model structure configuration}
  \label{tab:notation_model}
  \begin{center}
  \begin{tabular}{|l|c|l|c|}
    \hline  
    \textbf{Configuration}&\textbf{Value}&\textbf{Parameter}&\textbf{Value}\\\hline
    Input history address & $T_I$ & Feed forward dimension & $D_F$\\\hline
    Transformer input patches & $T_{T}$ &Output delta bitmap size & $D_O$\\\hline
    Input address dimension & $D_I$ & Transformer heads & $H$ \\\hline
    Attention dimension& $D_A$ & Transformer layer & $L$ \\\hline
\end{tabular}
\end{center}
\end{table}

\begin{table}[h]
  \caption{Notations for the table configuration shown in the format of $\langle$prototypes $K$, subspaces $C\rangle$}
  \label{tab:notation_tab}
  \begin{center}
  \begin{tabular}{|l|c|l|c|}
    \hline  
    \textbf{Layer}&\textbf{Configuration}&\textbf{Layer}&\textbf{Configuration}\\\hline
    Input Linear & $\langle K_I, C_I \rangle$ & Attention & $\langle K_A, C_A\rangle$\\\hline
    Feed Forward & $\langle K_F,C_F\rangle$  &Output Linear & $\langle K_O,C_O\rangle$  \\\hline
\end{tabular}
\end{center}
\end{table}

\subsubsection{Entire Model Latency and Storage Cost}
\label{sec:model_complexity}
For a network architecture in Figure~\ref{fig:vit_map}, we use notations for the model configuration as Table~\ref{tab:notation_model} and notations for the table configuration in~\ref{tab:notation_tab}. The tabularized entire model latency is shown in Equation~\ref{eq:model_latency} and the tabularized entire model storage is shown in Equation~\ref{eq:model_storage}, where $\mathcal{L}_{ln}$ and $\mathcal{S}_{ln}$ are the latency and storage for a Layer Normalization operation, $\mathcal{L}_{\sigma}$ and $\mathcal{S}_{\sigma}$ are for output Sigmoid activation function latency and storage cost.
\begin{equation}
\label{eq:model_latency}
\begin{aligned}
\mathcal{L}& = \underbrace{\mathcal{L}_l(K_I, C_I)}_{\text{Input linear}} +\underbrace{\mathcal{L}_{ln} +\mathcal{L}_l(K_O, C_O)+\mathcal{L}_{\sigma}}_{\text{Output linear}} \\
&+ \underbrace{L[\underbrace{ 2\mathcal{L}_{ln} + 2\mathcal{L}_l(K_A,C_A) +  \mathcal{L}_a(K_A,C_A)}_{\text{Multi-head self-attention}}+{\underbrace{2\mathcal{L}_l(K_F,C_F) }_{\text{Feed forward}}} ]}_{\text{Transformer encoder layers}}
\end{aligned}
\end{equation}
\begin{equation}
\label{eq:model_storage}
\begin{aligned}
\mathcal{S}& = \underbrace{2\mathcal{S}_l(T_I,D_A, K_I, C_I)}_{\text{Input linear}}
+\underbrace{\mathcal{S}_{ln} +\mathcal{S}_l(T_T,D_O, K_O, C_O)+\mathcal{S}_{\sigma}}_{\text{Output linear}} \\
&+ L[2\mathcal{S}_{ln} + \mathcal{S}_l(T_T, 3HD_A, K_A,C_A) \\  
&~~~~~~\underbrace{+ \mathcal{S}_a(T_T, D_A, K_A,C_A) + \mathcal{S}_l(T_T, D_A, K_A,C_A)}_{\text{Multi-head self-attention}}\\
&\underbrace{~~~~~+\underbrace{\mathcal{S}_{ln}+\mathcal{S}_l(T_T, D_F, K_F,C_F) + \mathcal{S}_l(T_T, D_A, K_F,C_F)]}_{\text{Feed forward}}}_{\text{Transformer encoder layers}}\\
\end{aligned}
\end{equation}
\subsubsection{Table Configurator} We pre-define a list of designs for the model configuration variables ($D_I, D_A, D_F,D_O,H,L$) and table configuration variables ($K,C$). Based on Equation~\ref{eq:model_latency} and Equation~\ref{eq:model_storage}, we generate a configuration dictionary that maps a specific configuration to its latency and storage cost. Based on the dictionary, we design a table configurator that uses a latency-major greedy approach to provide a valid configuration. It first finds configurations with the highest latency smaller than $\tau$. Under this latency, it finds the configuration with maximum storage smaller than $s$. If no such configuration exists, it moves to the lower latency configurations and continues the process, until the configuration meets both the latency and storage constraints.

\subsection{Complexity Reduction via Knowledge Distillation}
\label{sec:approach_kd}

The table configurator identifies valid model configurations, adhering to design constraints, which often result in smaller models than our trained large ones (Section~\ref{sec:approach_attn}).
We use Knowledge Distillation (KD)~\cite{hinton2015distilling} to transfer knowledge from the larger teacher model to the compact student model. 
Given that our task is multi-label classification, we utilize BCE loss and introduce a T-Sigmoid function (Equation~\ref{eq:sigmoid})—a softened variant inspired by T-Softmax~\cite{hinton2015distilling}—to refine probability distributions over multiple class outputs.
\begin{equation}
\label{eq:sigmoid}
z_i=p(y_i)_{t=T}=\sigma\left(\frac{y_i}{T}\right)=\frac{1}{1+e^{-y_i/T}}
\end{equation}
The complete loss function ($Loss$) encompasses both the BCE loss ($Loss_{BCE}$) and the soft KD loss ($Loss_{KD}$), as defined in Equation~\ref{eq:kd_loss},
\begin{equation}
\begin{aligned}
\label{eq:kd_loss}
Loss_{\mathrm{KD}}=\sum_{k=1}^q &\operatorname{KL}\left(\left[z_i^{{tch}}, 1-z_i^{{tch}}\right] \|\left[z_i^{{stu}}, 1-z_i^{stu}\right]\right)\\
Loss&=\lambda Loss_{\text {KD}} + (1-\lambda)Loss_{\text {BCE}}
\end{aligned}
\end{equation}
where $\operatorname{KL}(\cdot \| \cdot )$ is the Kullback-Leibler divergence~\cite{joyce2011kullback}, $\lambda$ is a hyper-parameter tuning the weights of the two losses, $z_i^{{tch}}$ and $z_i^{{stu}}$ are T-Sigmoid output of teacher and student models.
% \begin{equation}
% \label{eq:total_loss}

% \end{equation}

\subsection{Layer-Wise Tabularization with Fine-Tuning}
\label{sec:finetune}

\subsubsection{Layer-Wise Tabularization}
We convert the compressed model from KD to a hierarchy of tables layer by layer. The table configuration is given by the table configurator. The tabularization process is shown in Algorithm~\ref{alg:tabularize}. For each neural network layer, we check the operation and use the corresponding tabularization kernel introduced in Section~\ref{sec:kernel}, including linear kernel (line 10) and attention kernel (line 13). The output Sigmoid activation function can be approximated by a fixed lookup table~\cite{meher2010optimized}. Layer Normalization process is dimension-wise simple arithmetic operation without matrix multiplication, so we directly use the original operation. We initialize the output hierarchy of tables as an empty sequence $\mathcal{T}$ and push each converted layer to $\mathcal{T}$.

\begin{algorithm}[t]
  \caption{Layer-Wise Tabularization with Fine-Tuning}
  \label{alg:tabularize}
  \begin{algorithmic}[1]

    \State \textbf{Input:} Trained $N$-layer model $\mathcal{M}$, Training input data $\mathcal{D}$
    \State \textbf{Initialize:} Model layer $i$ output $ L[i] \leftarrow \mathcal{M}[0:i](\mathcal{D})$ 
    \State \textbf{Initialize:} Table hierarchy $\mathcal{T}$, fine-tune epoch $E$
    \State \textbf{Initialize:} Configuration lists prototypes $K$, subspace $C$
    \For{$i$ \textbf{in} $0$ to $N-1$} \Comment{Layer-wise tabularization}
        \If{$\mathcal{M}[i]$ is a linear layer}
            \If {$i>0$}: \Comment{Layer fine-tuning}
                %\State  layer input $I[i] = \mathcal{T}(\mathcal{D})$
                \State  $\mathcal{M}'[i] \leftarrow \Call{FineTune}{\mathcal{M}[i],\mathcal{T}(\mathcal{D}), L[i], E}$ 
            \EndIf
             \State $\mathcal{T}_{\text{linear}}  \leftarrow \Call{LinearKernel}{\mathcal{M}'[i], K[i], C[i]}$ 
             \State Push $\mathcal{T}_{\text{linear}}$ to $\mathcal{T}$
           % $\mathcal{T}_{linear} = \text{Linear_kernel}(\mathcal{M}'[i],K[i],C[i]$)
            %\State $\mathcal{M}'[i] \leftarrow$ Fine_tune $(\mathcal{M}[i])$
            %($M[i]$,$\mathcal{M}[0:i](\mathcal{D})$, $L[i]$)
        \ElsIf {$\mathcal{M}[i]$ is an attention layer}  
            \State $\mathcal{T}_{\text{attn}}  \leftarrow \Call{AttentionKernel}{\mathcal{M}[i], K[i], C[i]}$
             \State Push $\mathcal{T}_{\text{attn}}$ to $\mathcal{T}$
        \ElsIf {$\mathcal{M}[i]$ is a Sigmoid}
            \State Push Sigmoid lookup table approximation to $\mathcal{T}$
        \ElsIf {$\mathcal{M}[i]$ is a LayerNorm}
            \State Push LayerNorm parameters and operations to $\mathcal{T}$
        \EndIf
    \EndFor
    \State \textbf{Output:} $\mathcal{T}$
   
  \end{algorithmic} 
\end{algorithm}

\subsubsection{Fine-Tuning}
For a hidden linear layer $\bm{Y} = \bm{W}\bm{X} + \bm{B}$ (Equation~\ref{eq:linear}) of model $\mathcal{M}$, denoted as $\mathcal{M}[i]$, where $i>0$ is the layer index, we fine-tune the layer weights to achieve higher approximation performance (Algorithm~\ref{alg:tabularize} line 8). 
Assuming we have tabluarized layers prior to $\mathcal{M}[i]$, the input of $\mathcal{M}[i]$, which is the output of $\mathcal{M}[i-1]$, is now an approximation $\hat{\bm{X}}$ = $\bm{X} + \epsilon_x$ with the error $\epsilon_x$ introduced by tabularization. This leads to the error in the tabularized layer output $\bm{W}\hat{\bm{X}} + \bm{b}=\bm{Y} + \epsilon_y$. With an increasing number of layers being tarbularized, the errors accumulate, resulting in low approximation performance. To ameliorate the error accumulation problem, we fine-tune the layer weight $\bm{W}$ and bias $\bm{b}$. We initialize $\bm{W}$ and $\bm{b}$ as in the trained model, use the tabularization approximated $\hat{\bm{X}}$ as input and the original layer output $\bm{Y}$ as target, and performs $E$ epochs of layer training. We use the Mean Squared Error (MSE) loss function to learn the updated layer $\mathcal{M}'[i]$ with new weights $\bm{W'}$ and $\bm{b'}$:
\begin{equation}
    %\text{Loss} = \frac{1}{n} \sum_{i=1}^n \sum_{j=1}^d\left(\bm{Y}_{i j}-\hat{\bm{Y}}_{i j}\right)^2
    %\text{Loss} = \frac{1}{n} \sum_{i=1}^n \sum_{j=1}^d\left(\bm{Y}_{i j}-({\bm{W}\hat{\bm{X}_{i j}} + \bm{b}})\right)^2
    (\bm{W'},\bm{b'}) = \underset{\bm{W},\bm{b}}{\arg \min }\frac{1}{n} \sum_{i=1}^n \sum_{j=1}^d\left(\bm{Y}_{i j}-({\bm{W}\hat{\bm{X}_{i j}} + \bm{b}})\right)^2
%    T(\mathcal{M'(\mathbf{x}; \boldsymbol{\theta'})})<\tau
\end{equation}

After fine-tuning, we perform tabularization based on the updated layer $\mathcal{M}'[i]$. In this way, we train a table that imitates the output of a NN layer, going beyond mere approximation of the model weights. This novel approach draws inspiration from KD, framing the process as distilling knowledge from a NN layer into a table.

\section{Evaluation}

\subsection{Experimental Setup}

\subsubsection{Simulator}

Consistent with existing literature, we use ChampSim~\cite{gober2022championship} to generate traces and assess our approach. ChampSim simulates a modern heterogeneous multi-core system with an arbitrary memory hierarchy. The simulation parameters are detailed in Table \ref{tab:sim_param}, with prefetchers simulated at the last-level cache (LLC).
\begin{table}[h!]
\centering
\caption{Simulation parameters}
\setlength{\tabcolsep}{10pt} 
  \begin{tabular}{|c|c|}
    \hline
    \textbf{Parameter}&\textbf{Value}\\
   % \hline
    \hline
    CPU & 4 GHz, 4 cores, 4-wide OoO, \\
      & 256-entry ROB, 64-entry LSQ\\
    \hline
    L1 I-cache & 64 KB, 8-way, 8-entry MSHR, 4-cycle\\
     \hline
    L1 D-cache & 64 KB, 12-way, 16-entry MSHR, 5-cycle\\
     \hline
    L2 Cache&	1 MB, 8-way, 32-entry MSHR, 10-cycle\\
     \hline
    LL Cache&	8 MB, 16-way, 64-entry MSHR, 20-cycle\\
     \hline
     &	 $t_{RP}=t_{RCD}=t_{CAS} = 12.5$ ns\\
DRAM& 2 channels, 8 ranks, 8 banks\\ 
& 32K rows, 8GB/s bandwidth per core\\
  \hline
\end{tabular}
\label{tab:sim_param}
\end{table}

\subsubsection{Benchmark}

We evaluate \ourwork~on widely-used benchmarks: \emph{SPEC CPU 2006}~\cite{jaleel2010memory} and \emph{SPEC CPU 2017}~\cite{SPEC2017}. 
Table~\ref{tab:benchmarks} details the workloads we use and the LLC trace statistics extracted using ChampSim. The workloads consist of a diverse number of deltas and pages, which are representative to various scenarios.

\begin{table}[h!]
  \centering
  \caption{Benchmark application memory trace statistics}
  \begin{tabular}{|c|c|c|c|c|}
    \hline
    \textbf{Benchmark} &\textbf{Application} & \textbf{\# Address} & \textbf{\# Page}& \textbf{\# Delta} \\
    \hline
     & 410.bwaves & 236.5K& 3.7K& 14.4K \\
     \cline{2-5}
    SPEC 2006 &433.milc & 170.7K& 19.8K& 15.8K \\
    \cline{2-5}
    &437.leslie3d&104.3K&1.7K&3.6K \\
    \cline{2-5}
    &462.libquantum& 347.8K & 5.4K&0.5K \\
    \hline
    & 602.gcc & 195.8K& 3.4K& 4.9K \\
     \cline{2-5}
    SPEC 2017 & 605.mcf & 176.0K& 3.7K& 207.7K \\
    \cline{2-5}
    &619.lbm & 121.8K& 1.9K&1.2K \\
    \cline{2-5}
    &621.wrf&188.5K&3.3K&13.7K \\
    \hline
  \end{tabular}
  \label{tab:benchmarks}
\end{table}

\begin{table}[h!]
    \centering
    \caption{Configurations of models}
    \setlength{\tabcolsep}{4pt} 
    \begin{tabular}{|c|c|c|c|c|c|c|c|c|}
        \hline
    &\multicolumn{3}{c|}{\textbf{NN Config}}&\multicolumn{2}{c|}{\textbf{Table Config}}&\multicolumn{3}{c|}{\textbf{Complexity} }\\
    \cline{2-9}
     &\textbf{$L$} & \textbf{$D$} & \textbf{$H$} & \textbf{$K$}& \textbf{$C$} &\textbf{($\mathcal{L}$/cycle)}&\textbf{($\mathcal{S}$/B)}&\textbf{($\mathcal{A}$)} \\
     \hline
    %\textbf{Model} &\textbf{L} & \textbf{D} & \textbf{H}& \textbf{K} & \textbf{C} & \textbf{Size} \\
    Teacher&4&256&8& - & - &16.5K& 86.2M &98.3M\\
    \hline
    Student&1&32&2& - & - &908& 827.4K &134.7K\\
    \hline
    DART&1&32&2& 128& 2 & 97&864.4K &11.0K\\
    \hline
    \end{tabular}
    \label{tab:model_config}
\end{table}

\subsubsection{Model Configurations}
Table~\ref{tab:model_config} outlines configurations for the large Teacher model, compressed Student model, and an exemplar table-based predictor for~\ourwork. All models are implemented under full parallelism, the complexity of~\ourwork~is based on Section~\ref{sec:kernel_complexity}, whereas the NN-based models Teacher and Student is examined under systolic array implementation for matrix multiplications~\cite{kung1979systolic}.
We tailor the NN's layers ($L$), hidden dimensions ($D$), and heads ($H$). For tabularized models, consistent number of prototypes ($K$) and subspaces ($C$) are set across operations.

From our configuration, we measure latency $\mathcal{L}$, storage $\mathcal{S}$, and operations $\mathcal{A}$. Against the Teacher, \ourwork~achieves $170\times$ acceleration, $102\times$ compression, and a $99.99\%$ reduction in operations. Compared to the Student of similar size, \ourwork~offers $9.36\times$ acceleration, cutting operations by $91.83\%$.

\subsubsection{Metrics}

We evaluate multi-label memory access prediction using F1-score~\cite{goutte2005probabilistic}.
We evaluate the prefetching performance through simulation on ChampSim. In the simulation runs, we record the cache performance and report the prefetch accuracy, coverage, and IPC improvement~\cite{srinivasan2004prefetch}. The running time of the workloads on ChampSim cannot indicate the direct application performance. Instead, the IPC performance is reported based on the number of instructions and the cycles of simulation, which accurately indicates the prefetcher performance.

\subsection{Memory Access Patterns}
Figure~\ref{fig:patterns} shows the diverse memory access patterns of the benchmark applications in three scaled dimensions: instruction ID, page address, and block delta of consecutive accesses. 
The performance of the original large attention-based memory access prediction model is shown in Table~\ref{tab:exp_kd} as teacher models. The distribution of memory access patterns impacts the model prediction. We observe that applications showing fewer deltas are easier to predict compared with applications with more deltas under a similar number of pages, e.g., 602.gcc vs 605.mcf. We also observe that under a comparable number of deltas, applications with a smaller number of pages are easier to predict, e.g., 410.bwaves vs 433.milc.

\begin{figure}[ht]
    \centering
    \includegraphics[width=1\linewidth]{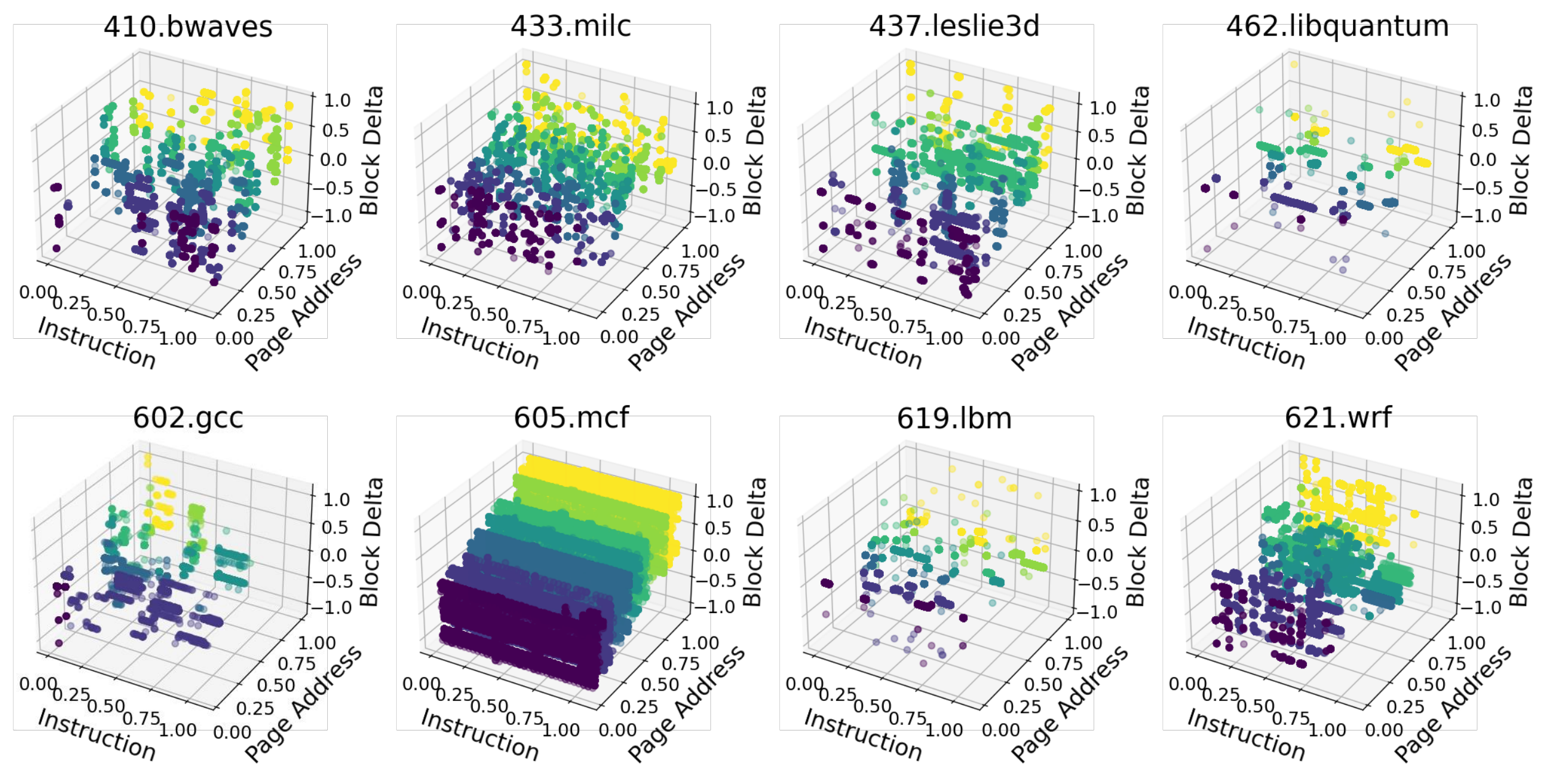 }
    \caption{{\color{black}Visualization of memory access patterns in the benchmark applications. The patterns show diversity in instructions, deltas, and pages, representing various scenarios.
    %The patterns are shown in three scaled dimensions: instruction ID, page address, and block delta of consecutive accesses. 
    }}
    %F1-scores of DART by varying the number of prototypes $K$.}
    \label{fig:patterns}
\end{figure}

\subsection{Evaluation of Multi-Label Knowledge Distillation}
Table~\ref{tab:exp_kd} shows the F1-score of the teacher model, the student model trained without knowledge distillation (Stu w/o KD), and the student model trained using the proposed multi-label knowledge distillation with T-Sigmoid activation function (Student). Results show that the compressed models trained using KD raise the mean F1-score from 0.751 to 0.783, which is only 0.005 lower than the 107$\times$ larger teacher model.

\begin{table}[ht!]
\centering
\caption{F1-score of the teacher model and the student models trained with and without knowledge distillation}
\setlength{\tabcolsep}{2pt} 
  \begin{tabular}{|c|c|c|c|c|c|c|c|c|c|c|}
  \hline
      &\multicolumn{8}{c|}{\textbf{Applications}}&\\ 
    \cline{2-9}

         \textbf{Models}&{\textbf{410.}}&{\textbf{433.}}&{\textbf{437.}}&{\textbf{462.}}&{\textbf{602}}&{\textbf{605.}}&{\textbf{619.}}&{\textbf{621.}}&{\textbf{Mean}}\\   
         \textbf{}&{\textbf{bwav}}&{\textbf{milc}}&{\textbf{lesl}}&{\textbf{libq}}&{\textbf{gcc}}&{\textbf{mcf}}&{\textbf{lbm}}&{\textbf{wrf}} &\\   
    \hline
    Teacher & 0.969&0.863&0.599&0.992&0.952&0.551&0.742&0.638&0.788\\
    \hline 
    Stu w/o KD &0.923&0.715&0.545&0.991&0.946&0.545&0.679&0.660&0.751\\
    \hline
    Student &\textbf{0.923}&\textbf{0.789}&\textbf{0.552}&\textbf{0.991}&\textbf{0.947}&\textbf{0.655}&\textbf{0.751}&\textbf{0.660}&\textbf{0.783}\\
\hline
\end{tabular}
\label{tab:exp_kd}
\end{table}

\subsection{Evaluation of Tabularization}

We assess layer-wise tabularization performance under varying table configurations, such as the number of prototypes ($K$) and subspaces ($C$), without fine-tuning. Our model structure remains fixed as the DART model (see Table~\ref{tab:model_config}), with the only variation being either $K$ or $C$. Figures~\ref{fig:f1_k} and \ref{fig:f1_c} illustrate how adjusting $K$ and $C$ impacts DART's F1-score. The results reveal that increasing $K$ significantly boosts prediction performance when $K$ exceeds 128, with $K=1024$ achieving a 10.9\% higher F1 score compared to $K=16$. In contrast, the impact of higher $C$ on F1-score is less pronounced, with $C=8$ outperforming $C=1$ by 6.6\%. While higher values of both $K$ and $C$ generally enhance tabularization performance, it's important to note that this comes at a substantial increase in latency and storage costs. Figure~\ref{fig:ls_kc} demonstrates a linear relationship between model latency and $\log(K)$ as well as $\log(C)$, while storage costs exhibit exponential growth.

\begin{figure}[h!]
    \centering
    \includegraphics[width=1\linewidth]{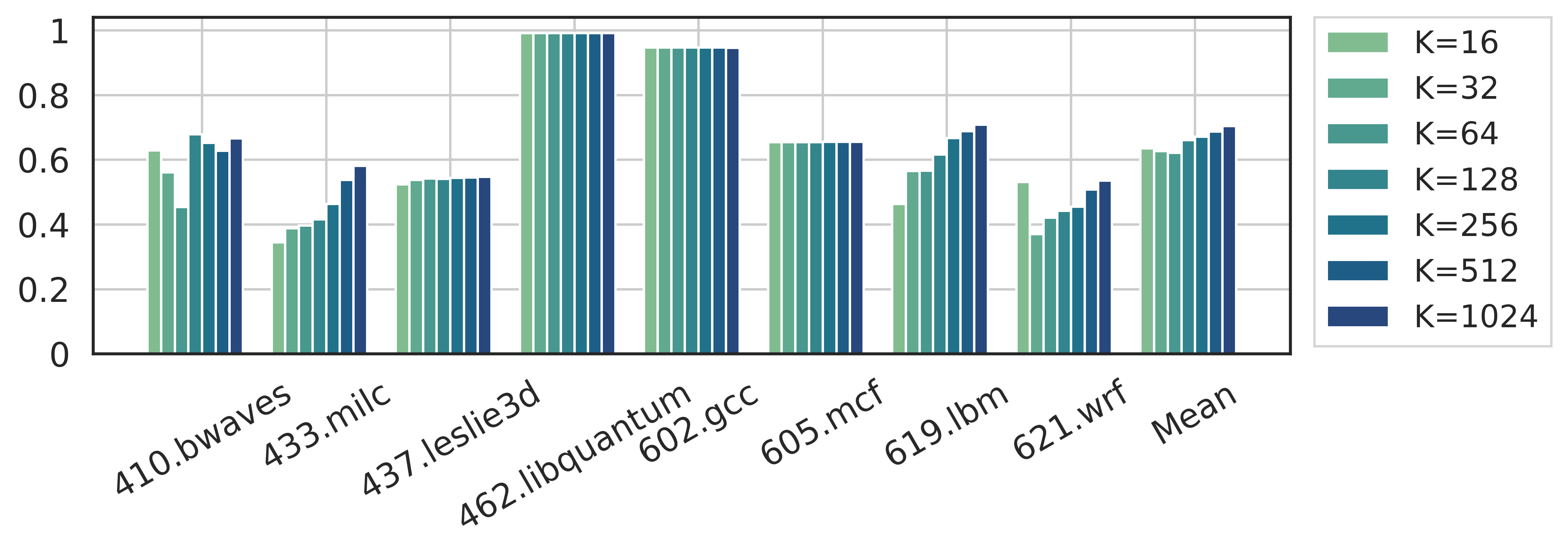 }
    \caption{F1-scores with varying DART prototypes $K$.}
    %F1-scores of DART by varying the number of prototypes $K$.}
    \label{fig:f1_k}
\end{figure}

\begin{figure}[h!]
    \centering
    \includegraphics[width=1\linewidth]{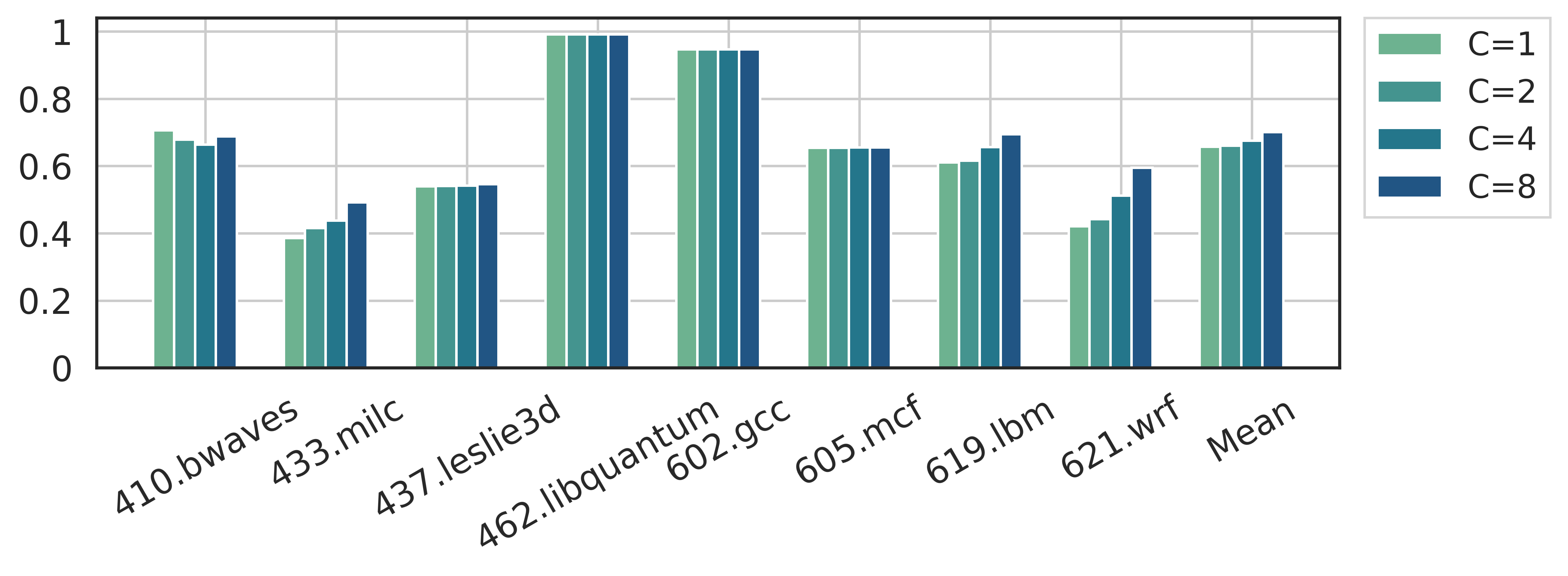 }
    \caption{F1-scores with varying DART subspaces $C$.}
    \label{fig:f1_c}
\end{figure}

\begin{figure}[h!]
    \centering
    \includegraphics[width=1\linewidth]{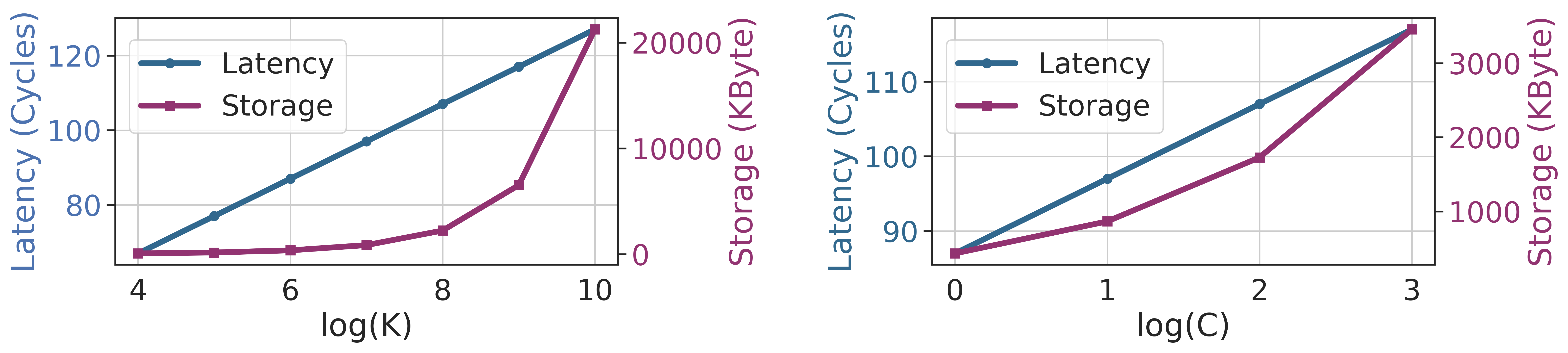 }
    \caption{Latency and storage cost under various $K$ and $C$. Model latency scales linearly with $\log(K)$ and $\log(C)$, whereas storage experiences exponential growth.}
    \label{fig:ls_kc}
\end{figure}

\subsection{Evaluation of Fine-Tuning}
We compare cosine similarity between Student NN and its tabularized models: DART with fine-tuning vs. DART without (DART w/o FT). Figure~\ref{fig:cossim} shows that fine-tuning effectively raises the layer cosine similarity, especially for the layers close to the output. Table~\ref{tab:dart_f1} shows the F1-score of the above models. On average, DART achieves an F1-score of 0.699, representing a 5.75\% gain over DART w/o FT but 0.084 lower than the Student model.
\begin{figure}[h!]
    \centering
    \includegraphics[width=1\linewidth]{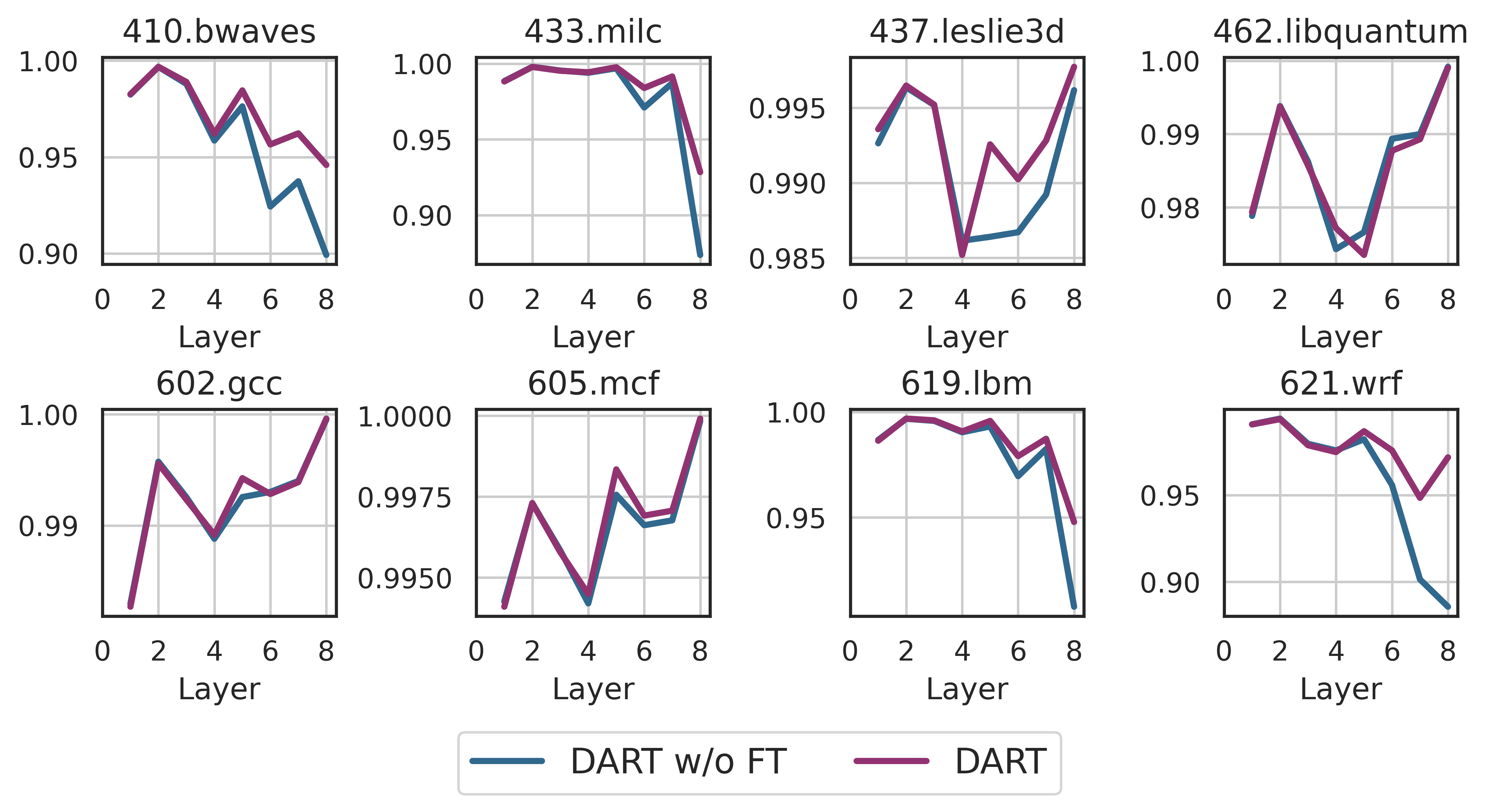 }
    \caption{Layer-wise cosine similarity comparison between DART without fine-tuning (DART w/o FT) and DART.}
    \label{fig:cossim}
\end{figure}

\begin{table}[ht!]
\centering
\caption{F1-score of DART for memory access prediction}
\setlength{\tabcolsep}{2pt} 
  \begin{tabular}{|c|c|c|c|c|c|c|c|c|c|c|}
  \hline
      &\multicolumn{8}{c|}{\textbf{Applications}}&\\ 
    \cline{2-9}

         \textbf{Models}&{\textbf{410.}}&{\textbf{433.}}&{\textbf{437.}}&{\textbf{462.}}&{\textbf{602}}&{\textbf{605.}}&{\textbf{619.}}&{\textbf{621.}}&{\textbf{Mean}}\\   
         \textbf{}&{\textbf{bwav}}&{\textbf{milc}}&{\textbf{lesl}}&{\textbf{libq}}&{\textbf{gcc}}&{\textbf{mcf}}&{\textbf{lbm}}&{\textbf{wrf}} &\\   
    \hline
 
    DART w/o FT &0.679&0.416&0.541&0.991&0.946&0.655&0.617&0.443&0.661\\
    \hline
    DART &\textbf{0.790}&\textbf{0.480}&\textbf{0.544}&\textbf{0.991}&\textbf{0.947}&\textbf{0.655}&\textbf{0.638}&\textbf{0.543}&\textbf{0.699}\\
\hline
\end{tabular}
\label{tab:dart_f1}
\end{table}

\subsection{Evaluation of Prefetching}

\subsubsection{Prefetcher Configurations}
As shown in Table~\ref{tab:3_darts}, we generate two more valid configurations using our table configurator given varied design constraints.
We evaluate the prefetching performance of \ourwork, along with small and large variants: DART-S and DART-L.

\begin{table}[h!]
    \centering
    \caption{Configurations of DART under various prefetcher design constraints}
    \setlength{\tabcolsep}{2.5pt} 
    \begin{tabular}{|c|c|c|c|c|c|}
        \hline
    \textbf{Prefetcher} & \textbf{Constraints} & \textbf{Configuration} &\textbf{Latency}&\textbf{Storage}&\textbf{Ops} \\
     & \textbf{$(\bm{\tau\text{/cycle}, s\text{/B}})$} & \textbf{$(L, D, H, K, C)$}&\textbf{($\mathcal{L}$/cycle)}&\textbf{($\mathcal{S}$/B)}&\textbf{($\mathcal{A}$)}\\
     \hline
    DART-S& 60, 30K & $1, 16, 2, 16, 1$&57 &29.9K&1.6K\\
    DART & 100, 1M & $1, 32, 2, 128, 2$ & 97 &864.4K& 11.0K \\
    DART-L&200, 4M & $2,32,2,256,2$ &191&3.75M&17.5K\\
    \hline
    \end{tabular}
    \label{tab:3_darts}
\end{table}

\subsubsection{Baseline Prefetchers}

Table~\ref{tab:baseline_pref} show all the prefetchers we implemented for evaluation. There are three types of baseline prefetchers: 1) practical rule-based prefetchers, including BO~\cite{michaud2016best} and ISB~\cite{jain2013linearizing}, 2) neural network based prefetchers, including TransFetch~\cite{zhang2022fine} and Voyager~\cite{shi2021hierarchical}, and 3) idealized versions of TransFetch and Voyager, which disregard practical hardware implementation constraints and assume zero latency in ChampSim simulation.

\begin{table}[h!]
    \centering
    \caption{Configurations of prefetchers for evaluation}
    \setlength{\tabcolsep}{3pt} 
    \begin{tabular}{|c|c|c|c|c|c|c|}
    \hline
    \textbf{Prefetcher} & \textbf{Storage} & \textbf{Latency} & \textbf{Table} & \textbf{ML} & \textbf{Mechanism}\\
    \hline
    BO~\cite{michaud2016best} & 4KB & $\approx$ 60 & $\checkmark$&& Spatial locality\\
    ISB~\cite{jain2013linearizing} & 8KB & $\approx$ 30 & $\checkmark$&& Temporal locality\\\hline
    TransFetch~\cite{zhang2022fine} &13.8MB&4.5K  &  &$\checkmark$& Attention\\
    Voyager~\cite{shi2021hierarchical} &14.9MB & 27.7K & &$\checkmark$& LSTM\\\hline
    TransFetch-I & - & 0  &  &$\checkmark$& Attention (Ideal)\\
    Voyager-I & - & 0 & &$\checkmark$& LSTM (Ideal)\\\hline
    \ourwork & 29.9K-3.75M & 57-191 & $\checkmark$&$\checkmark$& Attention\\
    \hline
    \end{tabular}
    \label{tab:baseline_pref}
\end{table}

\subsubsection{Prefetching Performance}

Figure~\ref{fig:pref_acc} shows the prefetch accuracy of \ourwork~and the baselines. DART-S, DART, and DART-L achieve prefetch accuracy at 80.6\%, 80.7\%, and 82.5\%, respectively. While TransFetch-I and Voyager-I achieve the two highest accuracy values at 89.6\% and 95.1\%, their performance drops significantly with latency introduced in simulation to 78.6\% and 49.9\%. BO as a rule-based prefetcher shows high prefetch accuracy at 89.4\%.

\begin{figure}[h!]
    \centering
    \includegraphics[width=1\linewidth]{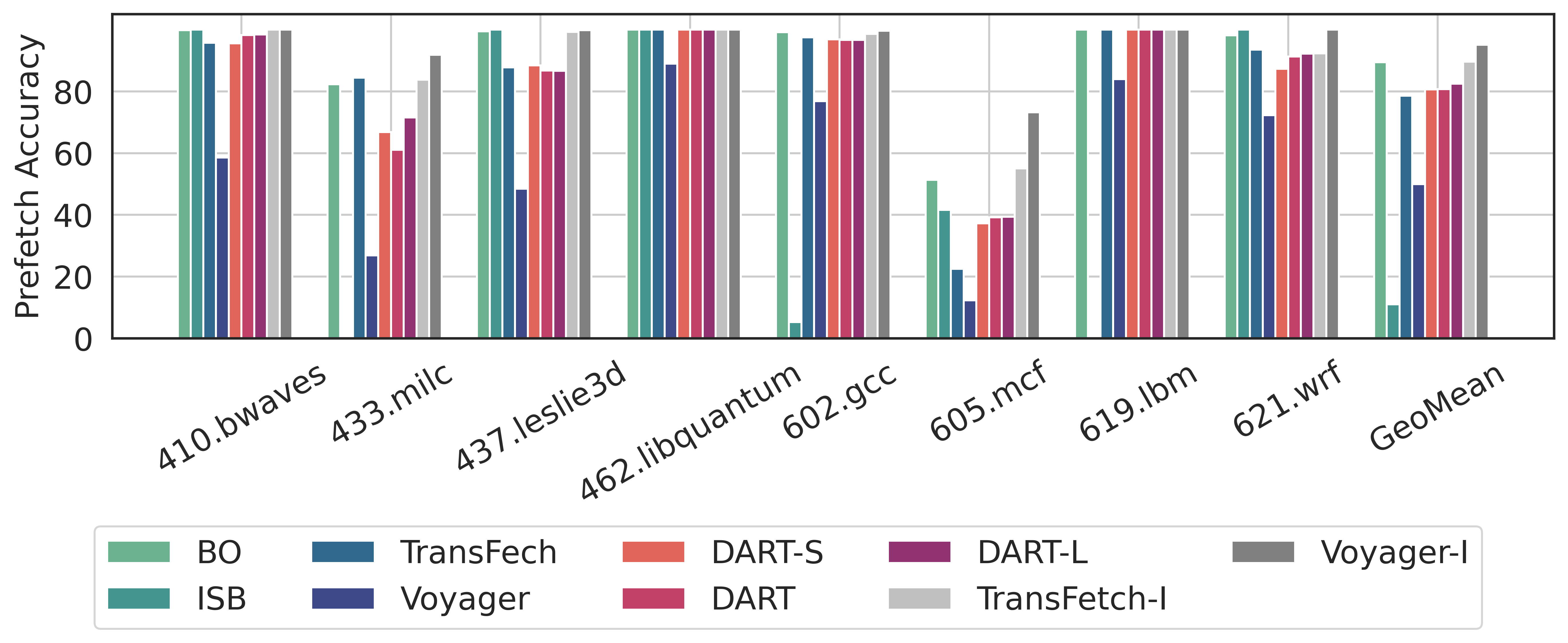 }
    \caption{Prefetch accuracy of~\ourwork~and the baselines.}
    \label{fig:pref_acc}
\end{figure}

\begin{figure}[h!]
    \centering
    \includegraphics[width=1\linewidth]{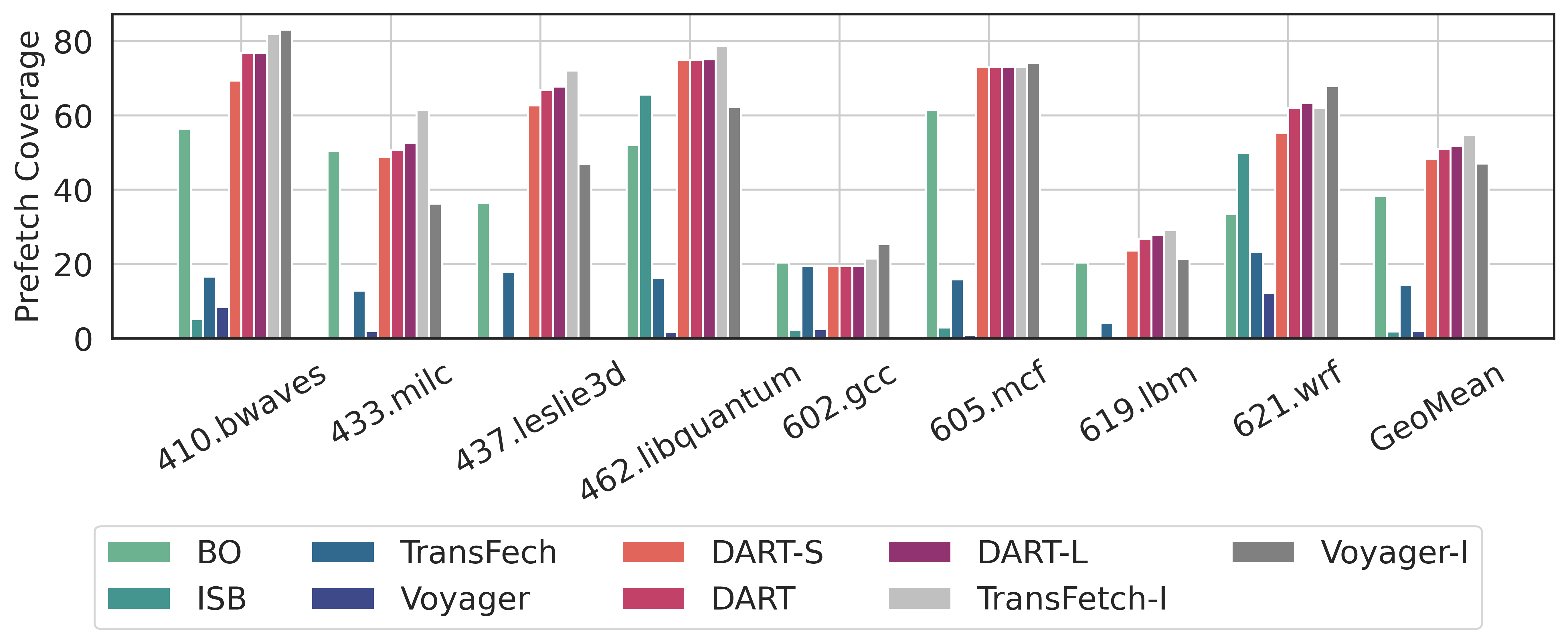 }
    \caption{Prefetch coverage of~\ourwork~and the baselines.}
    \label{fig:pref_cov}
\end{figure}
\begin{figure}[h!]
    \centering
    \includegraphics[width=1\linewidth]{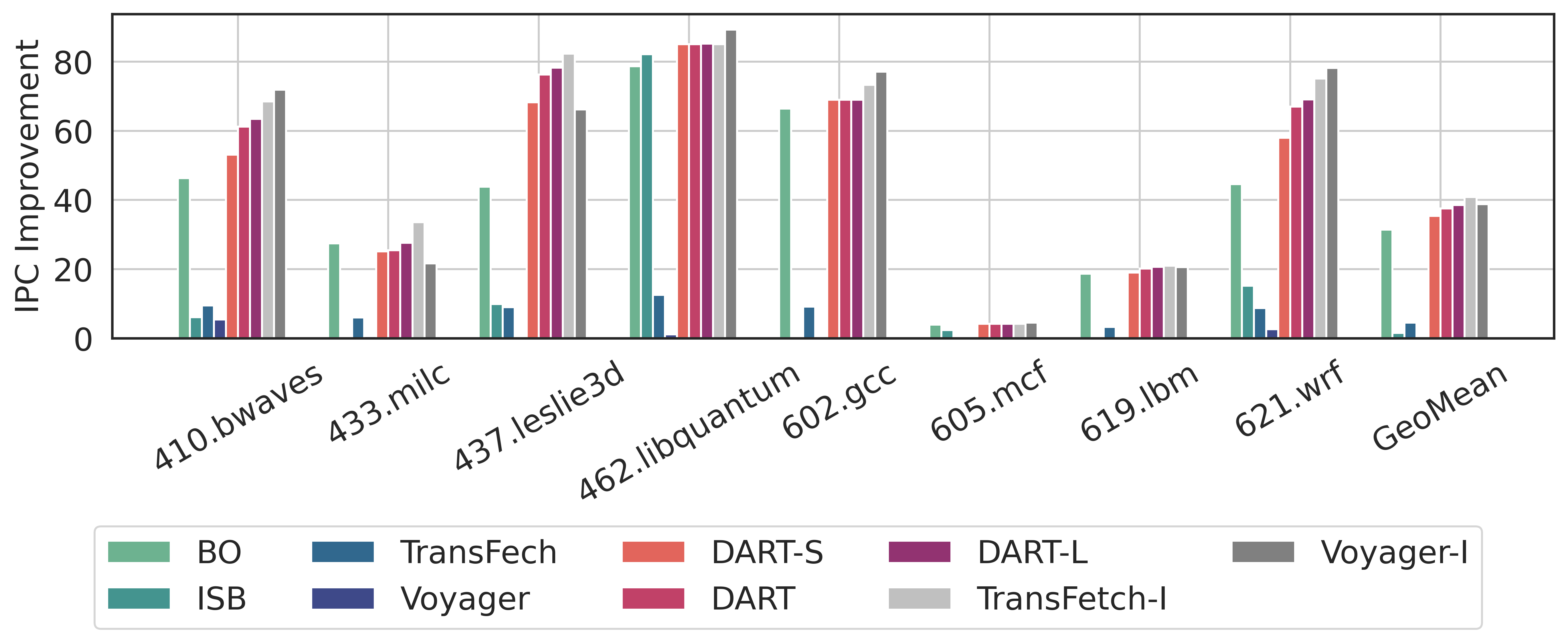 }
    \caption{IPC Improvement using~\ourwork~and the baselines.}
    \label{fig:pref_ipc}
\end{figure}

Figure~\ref{fig:pref_cov} shows the prefetch coverage of \ourwork~and the baselines. DART-S, DART, and DART-L achieve coverage at 48.3\%, 51.0\%, and 51.8\%, respectively. TransFetch-I and Voyager-I show coverage at 54.7\% and 47.0\%, but the values drop significantly to 14.4\% and 2.1\% with latency introduced. 

Figure~\ref{fig:pref_ipc} shows the overall IPC improvement when applying a prefetcher. DART-S, DART, and DART-L achieve IPC improvement at 35.4\%, 37.6\%, and 38.5\%, respectively.
All DART variants outperform the rule-based prefetchers BO (31.5\%) and ISB (1.6\%) and outperform the NN-based prefetchers TransFetch (4.5\%) and Voyager (0.38\%).
The basic configuration DART achieves 37.6\% IPC improvement, outperforming BO by 6.1\% and outperforming TransFetch by 33.1\%, only 3.3\% lower compared to the highest ideal and impractical prefetcher TransFetch-I at 40.9\%.
Even the smallest variant DART-S achieves 35.4\% IPC improvement under storage and latency comparable to rule-based prefetchers, outperforming all other baseline prefetchers besides the ideal NN-based prefetchers TransFetch-I and Voyager-I, which it underperforms by 5.5\% and 3.4\% respectively.

\section{Conclusion}

We presented DART, a prefetcher exemplifying our innovative approach, which distills knowledge from an attention-based neural network into a hierarchy of tables for practical NN-based data prefetching.
The keys to our approach include training an attention-based NN for memory access prediction, distilling this model into a compact form to meet design constraints, and implementing tabularization to create a hierarchy of tables.
DART maintains the prediction performance while significantly reducing the NN-based predictor's arithmetic operations and inference latency.
DART outperforms state-of-the-art table-based prefetchers under comparable storage and latency cost while achieving close performance to the impractical ML-based prefetchers w.r.t. IPC improvement.
In future work, we aim to optimize the tabularized model by reducing encoding and aggregation overheads.
Additionally, we plan to explore converting multiple layers into a single table to further reduce latency, storage, and operations, enhancing practicality towards real-world ML-based prefetching.

\section*{Acknowledgment}

This work has been supported by the U.S. National Science Foundation (NSF) under grant CNS-2009057 and SaTC-2104264, as well as the DEVCOM Army Research Lab (ARL) under grant W911NF2220159.

\textbf{Distribution Statement A}: Approved for public release. Distribution is unlimited.

\bibliographystyle{IEEEtran}
\bibliography{reference}

\end{document}